\documentclass[11pt]{article}

\usepackage[a4paper,margin=2.3cm]{geometry}
\usepackage{amsmath,amssymb,bm}
\usepackage{graphicx}
\usepackage{placeins}
\usepackage{booktabs}
\usepackage{hyperref}
\usepackage{xcolor}

\newcommand{\E}{\mathbb{E}}
\newcommand{\PP}{\mathbb{P}}
\newcommand{\MMD}{\mathrm{MMD}}
\newcommand{\IF}{\mathrm{IF}}
\newcommand{\AUC}{\mathrm{AUC}}
\newcommand{\Var}{\mathrm{Var}}

\title{
Localizing Global Discrepancies:
Marginal Contributions and Contextual Anomaly Detection
}

\author{
Tommaso Dorigo\\[0.5em]
\small INFN, Sezione di Padova, Padova, Italy;\\
\small Department of Computer Science, Electrical and Space Engineering,\\
\small Luleå University of Technology, Luleå, Sweden
}

\date{\today}

\begin{document}

\maketitle

\begin{abstract}

Global goodness-of-fit and discrepancy statistics can establish that a sample departs from a reference distribution without identifying which observations drive the departure.  We develop a unified framework for this localization problem by assigning to each observation its conditional or marginal contribution across random statistical contexts.  Marginal-attribution constructions are familiar from resampling diagnostics and modern data valuation; here we connect them systematically to projection theory and to
event-level anomaly detection.  For any symmetric statistic, fixed-size
replacement is exactly equivalent to centered conditional localization.  For U-statistics, the corresponding addition score equals the first Hoeffding/H\'ajek contribution; for smooth distributional functionals the score is related at leading order to the influence function; and for the unbiased known-background maximum mean discrepancy (MMD) it reduces exactly to the MMD witness.

The above viewpoint reveals how to construct more efficient estimators of the same local quantity. Matched-context subtraction removes fluctuations unrelated to the observation, while for pairwise MMD the event-containing terms yield a particularly simple localizer.
On the LHC Olympics benchmark, a particle physics anomaly detection dataset, the pair estimator converges to the direct empirical MMD witness with the predicted $1/(Rm^2)$ finite-resampling scaling, with $m$ the batch size and $R$ the number of batches. At $m=1000$ and $R=5\times10^6$, it reproduces the witness almost exactly, reaching correlation $0.9993$ and essentially identical AUC.

We separately study when the surrounding sample contains information about an event beyond that contained in its own features. 
To isolate this possibility, we construct a shared-latent toy model in which the full single-event distributions of signal and background are identical by construction, forcing every isolated-event classifier to have $\AUC=0.5$. Discriminating information survives only in the dependence between events induced by the latent parameter shared across signal observations.
In this setup, the ensemble reveals the common latent alternative and recovers class information, whereas an independent-latent control does not. This separates two roles of context: efficient localization of a global discrepancy, and genuinely additional class information when the alternative contains shared structure. The framework thereby connects classical attribution and projection ideas to event-level anomaly detection and clarifies what contextual methods can and cannot learn.

\end{abstract}

\section{Introduction}
\label{sec:introduction}

Searches for unexpected phenomena are often formulated as tests of whether an observed data sample is compatible with a reference probability distribution.  This setting is typical in high-energy physics (HEP). In experimental HEP analyses, Standard Model processes may be accurately simulated or otherwise constrained, thanks to well-developed theoretical calculations and high-fidelity simulations; the probability distribution of Standard Model data is not known in closed form, but can in principle be determined to high accuracy. The probability distribution associated with possible new physics is instead a priori completely unknown.  A wide class of modern anomaly-search methods therefore aims to identify departures of collider data from a well-specified reference without committing to a particular signal model \cite{dagnolo2019,hallin2022}.

A distinction needs to be made between identifying an anomalous
\emph{sample} and identifying the individual observations associated with the sample-level departure.  An event can lie in an ordinary region of background phase space while belonging to an anomalous population because too many events with similar properties occur in the data.  Conversely, an isolated event exhibiting extreme features need not support a coherent distributional discrepancy.  Thus
``atypicality under the background'' and ``contribution to the observed
departure'' are different statistical notions.

Assigning a global quantity back to individual observations is itself a
well-established statistical idea.  Case-deletion and jackknife methods study the sensitivity of statistics to individual data points, and the jackknife-after-bootstrap extracts such information from an existing ensemble of bootstrap replicates \cite{efron1992}.  Related constructions appear in data valuation, where Shapley-based and distributional approaches average the marginal effect of adding an observation over many subsets or random contexts
\cite{ghorbani2019,ghorbani2020,kwon2022,lin2022}.  Global-test diagnostics have likewise been used to expose observation- or variable-level contributions to an aggregate result \cite{goeman2004}, while for kernel two-sample tests the maximum mean-discrepancy (MMD) witness directly identifies regions in which two distributions disagree \cite{lloyd2015}. The mentioned research directions establish important pieces of the problem, but they arise in different settings and with different objectives.

The anomaly-detection motivation considered here has an independent history in the Inverse Bagging algorithm. That idea was conceived by the author in 1992 for the top quark search at the Tevatron collider, but was rapidly abandoned due to its impossible CPU demands. The algorithm was then studied more systematically 25 years later \cite{vischia2017}.  Inverse Bagging repeatedly constructs small subsets of a mixed data sample, evaluates their compatibility
with the background hypothesis, and attributes the resulting information back to the events appearing in those subsets.  The historical construction was phrased in terms of bootstrap aggregation. The perspective we develop here is broader: we wish to discuss how conditional subset scores, marginal contributions, statistical projections, and event-containing terms can be viewed as different representations of the same localization operation when the parent object is a global discrepancy statistic.

Let $T_m$ be a sample-level statistic computed on a batch of $m$ observations, assumed to be permutation-invariant (symmetric) in its arguments. Two natural questions can be asked about an observation $x$:
First, what value does $T_m$ tend to take when $x$ is present? And second, how much does $T_m$ tend to change because $x$ is present?
The first question leads to conditional localization,
\[
L_{T,m}(x)=\E[T_m\mid X_1=x]-\E[T_m],
\]
where $\E[]$ indicates the expectation value;
while the second leads to a marginal-contribution construction.  These are not competing definitions.  Replacing an ordinary member of a random context by $x$ gives exactly $L_{T,m}(x)$ for any symmetric statistic; for unbiased U-statistics, adding $x$ to a context of size $m-1$ gives the same quantity exactly.  The marginal form is also computationally important because matched-context subtraction cancels fluctuations that carry no information about the identity of $x$.

Starting from those questions and motivated by the attempt to understand under what conditions and to what extent the Inverse Bagging algorithm works, we studied the above landscape of ideas. The contribution we provide with this paper is threefold.  First, we give a unified statistical account of discrepancy localization: for
U-statistics the local score is the first Hoeffding/H\'ajek projection, for smooth functionals it connects at leading order to the influence function, and for known-background MMD it becomes the witness function.  These component results are classical or closely related to established constructions; here we stress their common interpretation as event-level localization of a global discrepancy.  Second, that interpretation exposes variance-efficient
estimators based on matched marginal contributions and, for U-statistics, on terms that explicitly contain the event being scored.  We derive their scaling and test it from analytically controlled examples to collider-scale data.  Third, we separate localization from the distinct question of whether other observations contain \emph{new class information} about a given event, and identify shared latent structure as a mechanism by which such information can arise.

Maximum mean discrepancy \cite{gretton2012} provides an especially transparent realization for our study, as we can derive all properties exactly.  Its unbiased known-background estimator is a second-order U-statistic, its conditional localization is exactly proportional to the MMD witness function, and its event-containing pair terms can be isolated analytically.  This allows a direct comparison between the population localizer, finite-sample witnesses, and Monte Carlo estimators, and makes the cancellation of context-only fluctuations explicit.


A separate question is whether the surrounding ensemble merely helps to estimate an event-level localizer, or whether it can contain genuinely new information about the class of a particular observation. 
In an ordinary IID mixture with fully specified signal and background distributions, once an event's own features are known, the remaining independent observations provide no additional information about its class. This limiting case is closely related to the analysis of single-event and multi-event collider classifiers in Ref.~\cite{nachman2021}.
The situation changes when the alternative itself contains an unknown property shared across signal events. The ensemble can then constrain that common property and use it to reinterpret individual observations. We construct an extreme example in which isolated signal and background events have exactly identical marginal distributions, yet their shared latent structure makes class membership recoverable from context.

The numerical studies we offer in this work follow the above conceptual hierarchy.  Experiment~I
provides an analytically soluble Gaussian validation and quantifies the variance reduction that can be obtained by marginal and event-containing estimators. Then, we perform
experiment~II where we show how to to isolate genuinely collective information with a shared latent alternative and an independent-latent control. In experiment~III we finally get to a HEP context; we decided to use the LHC
Olympics benchmark \cite{lhcolympics2021} in the standard CATHODE signal region to study convergence at collider-scale sample size.  
Writing $\rho$ for the correlation with the direct empirical witness, the pair estimator obeys $\rho^{-2}-1\simeq C(Rm^2)^{-\alpha}$ with $\alpha=0.9995$ in the high-statistics regime and recovers the direct empirical witness to correlation $0.9993$, with essentially identical AUC. Together, the three experiments distinguish validation of the localization principle, the existence of genuinely contextual class information, and the efficient recovery of the localizer in a HEP context.

The paper is organized as follows. Section~\ref{sec:related} positions the construction relative to resampling diagnostics, data valuation, projection theory, and anomaly detection. Section~\ref{sec:problem} introduces conditional and marginal localization. In Section~\ref{sec:theory} we develop the connections with U-statistics, influence functions, MMD, and efficient event-containing estimators. In Section~\ref{sec:information} we discuss how localization is separated from genuinely collective class information. Section~\ref{sec:algorithm} discusses finite-sample estimation, practical localization estimators, and the historical relation to Inverse Bagging. In Section~\ref{sec:toys} we present the three numerical studies mentioned above, and Section~\ref{sec:discussion} discusses their implications, computational consequences, and limitations. We conclude in Section 9 where we summarize the work and discuss its possible extensions.

\section{Related work}
\label{sec:related}

The construction studied here lies at the intersection of several mature branches of statistics literature.  Marginal changes under deletion or insertion, conditional averaging over subsets, low-order statistical projections, and pointwise kernel witnesses are all established objects.  The purpose of this section is to delimit the
specific synthesis developed in the remainder of the paper.

\subsection{Resampling diagnostics and observation influence}

Jackknife and case-deletion methods quantify how a statistic changes when an observation is removed from a realized sample. The technique of jackknife-after-bootstrap developed by Brian Efron showed that observation-level sensitivity can also be recovered from an already generated bootstrap ensemble, without a fresh resampling calculation for every point \cite{efron1992}.  Resampling has also
been used more directly for outlier localization: the multihalver of Fernholz, Morgenthaler and Tukey compares statistics on complementary half-samples and uses those differences to nominate observations with pronounced influence \cite{fernholz2004}.  In another specialized setting, diagnostics for the global test of Goeman et al. decompose an aggregate score into subject-level contributions that are proportional to the effect of deleting an individual subject \cite{goeman2004}.

The above mentioned constructions are close in spirit to the present problem, but their conditioning structure and objective differ.  Our target is the expected contribution of a fixed observation across random statistical contexts to a chosen global discrepancy from a reference distribution.  The matched-context formulation then uses paired evaluations not only as an influence diagnostic, but also as a variance-efficient Monte Carlo estimator of that population localizer.

\subsection{Data valuation and marginal effects over subsets}

The closest generic mathematical antecedents to our work arise in modern data valuation. Data Shapley averages the marginal effect of a training datum over coalitions of different sizes \cite{ghorbani2019}; Distributional Shapley extends the valuation to points considered relative to an underlying data distribution \cite{ghorbani2020}.  Beta Shapley makes explicit the role of fixed-cardinality marginal contributions and studies their statistical properties \cite{kwon2022}.  Particularly close to Eq.~\eqref{eq:addition-score}
discussed {\em infra} (Sec.\ref{sec:problem}) is the average marginal effect (AME) of Lin et al., defined as the expected
change in a utility when a point is added to a random subset drawn from a user-specified subset distribution \cite{lin2022}.

A complementary line estimates influence by repeatedly training on random subsets.  Feldman and Zhang introduced subsampled influence measures that compare model behavior across subsets containing or excluding particular training examples \cite{feldman2020}.  Data-OOB later showed that a bagging ensemble can itself be reused to obtain scalable pointwise data values, with a connection to infinitesimal-jackknife influence \cite{kwon2023}.

The mentioned results show that context-averaged marginal contributions, as well as the reuse of subset ensembles for pointwise attribution, are well-established constructions. In this work, we specialize these ideas to a different statistical object and interpretation. We take the value assigned to a subset to be a sample-level discrepancy from a reference distribution; we treat the context size $m$ as a parameter of the localization and estimation procedure rather than as part of an axiomatic coalition weighting; and we interpret the resulting event score as a contribution to the observed distributional departure. This specialization provides the setting in which the connections to classical projection theory developed below become particularly transparent.

\subsection{Projection theory and kernel discrepancy localization}

The projection results relevant to this construction belong to classical asymptotic statistics. For U-statistics, the Hoeffding decomposition separates a statistic into components of increasing interaction order, with the first-order H\'ajek projection isolating the contribution associated with individual observations \cite{hoeffding1948,serfling1980}. As shown in Sec.~\ref{sec:ustat}, the conditional localizer considered here coincides exactly with this first-order contribution, up to the conventional normalization.

Beyond U-statistics, von Mises and influence-function calculus describe the first-order response of a smooth statistical functional to an infinitesimal perturbation of the underlying distribution \cite{vonmises1947,hampel1974,vanderVaart1998}. In Sec.~\ref{sec:influence}, the same localization operation is shown to recover this first-order contribution asymptotically. Our use of both sets of results is therefore interpretive: they provide the classical statistical structures underlying the event-level localization considered here.

For MMD, pointwise localization is also well established in the literature.  The witness function of a kernel two-sample test identifies regions in which two distributions differ
\cite{gretton2012,lloyd2015}.  More recently, the KAIROS framework allows to formulate model-agnostic data valuation through the influence of an individual example on the MMD between an empirical training distribution and a clean reference, obtaining a
closed-form MMD influence score and a leave-one-out ranking
\cite{zhu2025}.  Accordingly, the MMD witness/influence identity in this work is not presented as a new kernel result.  Its role is to provide an exactly soluble member of the broader localization framework and to expose the finite-$m$ event-containing estimator whose Monte Carlo scaling can be tested directly.

\subsection{Anomaly detection and collective event information}

The direct anomaly-detection antecedent to the work we present in this manuscript is Inverse Bagging \cite{vischia2017}, which assigns small-batch goodness-of-fit information back
to the observations appearing in those batches.  The present analysis gives that procedure a statistical target and separates the raw conditional-average estimator from lower-variance marginal and event-containing estimators.

In high-energy physics, modern model-independent searches often learn a
density ratio or classifier that is capable of separating data from a reference background, as in NPLM-like approaches and CATHODE
\cite{dagnolo2019,hallin2022}\footnote{ An up-to-date living review of existing literature on anomaly detection is accessible at \url{https://iml-wg.github.io/HEPML-LivingReview/}}.  Other work has explicitly considered learning from multiple collider events at once and clarified when event ensembles can
or cannot improve classification \cite{nachman2021}.  Our emphasis is complementary. Given a global discrepancy, we first ask how it induces an event-level ranking and how that ranking can be estimated efficiently. We then address a separate question: whether the surrounding events can contain additional class information about a particular observation beyond what is available from that observation alone.

Taken together, the literature already contains the principal ingredients of
marginal attribution, influence, low-order projection, and MMD localization.
The contribution pursued here is their common formulation as localization of
a global discrepancy, the resulting efficient matched-context and
event-containing estimators, and the explicit separation between a
population-defined localizer and genuinely collective information generated
by shared latent structure.

\section{From global discrepancy to local contribution}
\label{sec:problem}

Consider observations in a mixture model,
\[
X_1,\ldots,X_N\sim P,
\qquad
P=(1-\epsilon)B+\epsilon S,
\]
where $B$ is a reference distribution and $S$ an unknown contaminating
component. Let $T_m(X_1,\ldots,X_m;B)$ be a symmetric sample-level
discrepancy statistic evaluated on batches of size $m$.

\subsection{Conditional localization}

For a fixed observation $x$, we first define the conditional average
\begin{equation}
G_{T,m}(x)
=
\E_P[T_m\mid X_1=x].
\label{eq:conditional-average}
\end{equation}
Its centered version,
\begin{equation}
L_{T,m}(x)
=
G_{T,m}(x)-\E_P[T_m]
=
\E_P[T_m\mid X_1=x]-\E_P[T_m],
\label{eq:localization}
\end{equation}
will be called conditional localizer in the following. Since the centering term does not depend on $x$, $G_{T,m}$ and $L_{T,m}$ induce the same event ranking.

Equation~\eqref{eq:localization} measures how the expected global
discrepancy changes when one member of the sample is fixed to $x$. It is
therefore a measure of the association of $x$ with the sample-level
departure, not a posterior probability that $x$ belongs to a signal class.

This distinction separates discrepancy localization from ordinary outlier
scoring. An observation can be improbable under $B$ yet contribute little
to a coherent departure of $P$ from $B$, while events in a modest but
localized excess can individually be quite probable under $B$. A
discrepancy localizer therefore ranks observations by their alignment with
the departure exhibited by the ensemble, rather than by their isolation
under the reference model.

\subsection{Marginal-contribution representations}
\label{sec:marginal-principle}

The same population localization can be expressed in terms of marginal
changes of the parent statistic. We use \emph{marginal contribution} as an
umbrella term for such matched differences, and distinguish two closely
related constructions.

\paragraph{Replacement marginal contribution.}
Let $C=(X_2,\ldots,X_m)$ be a random context and let $Y\sim P$ be
independent of $C$. Define the fixed-size replacement marginal contribution
by
\begin{equation}
A^{\rm rep}_{T,m}(x)
=
\E_{C,Y}\!
\left[
T_m(x,C)-T_m(Y,C)
\right].
\label{eq:replacement-score}
\end{equation}
For any symmetric statistic with finite expectation,
\begin{equation}
A^{\rm rep}_{T,m}(x)
=
L_{T,m}(x).
\label{eq:replacement-exact-localization}
\end{equation}
Thus the centered conditional localizer has an exact marginal-contribution
interpretation: it is the expected change in the statistic when an ordinary
member of a random batch is replaced by $x$.

\paragraph{Addition marginal contribution.}
A closely related construction compares a batch containing $x$ with the
same context before $x$ is added:
\begin{equation}
A^{\rm add}_{T,m}(x)
=
\E_C\!
\left[
T_m(x,C)-T_{m-1}(C)
\right].
\label{eq:addition-score}
\end{equation}
Its relation to the conditional localizer is
\begin{equation}
A^{\rm add}_{T,m}(x)
=
L_{T,m}(x)
+
\E[T_m]-\E[T_{m-1}].
\label{eq:addition-localization-relation}
\end{equation}
The difference is independent of $x$, so $A^{\rm add}_{T,m}$ and
$L_{T,m}$ always induce the same population ranking. For statistic
families that are unbiased for a common population target, including
U-statistics,
\[
\E[T_m]=\E[T_{m-1}],
\]
and therefore
\begin{equation}
A^{\rm add}_{T,m}(x)
=
L_{T,m}(x)
=
A^{\rm rep}_{T,m}(x).
\end{equation}

Conditional localization and marginal contribution are thus alternative
representations of the same event-level population quantity. The distinction
becomes important at the level of estimation: a raw Monte Carlo estimate of
the conditional average $G_{T,m}(x)$ and matched Monte Carlo estimates of
$A^{\rm rep}_{T,m}(x)$ or $A^{\rm add}_{T,m}(x)$ have the same population
ranking target, but can have very different variances.

\section{Statistical structure of local contribution}
\label{sec:theory}

\subsection{U-statistics and the H\'ajek projection}
\label{sec:ustat}

A U-statistic is obtained by applying a symmetric kernel of fixed order r to every distinct r-tuple of observations and averaging the results. It provides an unbiased estimator of the corresponding population expectation. Let
\begin{equation}
T_m
=
\binom{m}{r}^{-1}
\sum_{1\leq i_1<\cdots<i_r\leq m}
h(X_{i_1},\ldots,X_{i_r})
\label{eq:ustat-definition}
\end{equation}
be a U-statistic of order $r$, with
$\theta=\E[h(X_1,\ldots,X_r)]$, and
define the first Hoeffding projection
\[
h_1(x)
=
\E\!\left[h(x,X_2,\ldots,X_r)\right]-\theta.
\]

Conditioning on $X_1=x$ separates the kernel terms into the fraction
$r/m$ that contain the fixed observation and the fraction $1-r/m$ that do
not.  Hence, exactly for every $m\geq r$,
\begin{equation}
\boxed{
L_{T,m}(x)
=
\frac{r}{m}h_1(x).
}
\label{eq:ustat-main}
\end{equation}
This elementary identity is also an immediate consequence of the classical
Hoeffding decomposition \cite{hoeffding1948,serfling1980}.  In particular,
the first-order H\'ajek projection satisfies
\begin{equation}
\Pi_{\rm H}(T_m-\theta)
=
\frac{r}{m}\sum_{i=1}^m h_1(X_i)
=
\sum_{i=1}^m L_{T,m}(X_i).
\label{eq:hajek-sum-localizations}
\end{equation}
Thus contextual averaging isolates, observation by observation, the
first-order component of the global U-statistic.  The projection result is
classical; our use of it is as an event-level localization of a sample-level
discrepancy.

Two consequences of Eq.\eqref{eq:ustat-main} will be useful below. First, for a fixed parent
distribution and kernel, changing the batch size $m$ only rescales the
population localizer by the positive factor $r/m$; the induced event ordering is therefore exactly independent of $m$. Second, if $h_1 \equiv 0$, the
U-statistic is first-order degenerate and single-observation localization vanishes, although higher-order components of the statistic may remain. In
that case, localization must be formulated at the level of pairs or
higher-order groups of observations.

\subsection{Smooth distributional functionals and influence functions}
\label{sec:influence}

The exact first-order localization obtained for U-statistics has an asymptotic analogue for more general smooth distributional functionals. Let a population discrepancy be represented by a sufficiently regular functional 
$D(P,B)$.  
Under the standard infinitesimal contamination path
\[
P_{\eta,x}=(1-\eta)P+\eta\delta_x,
\]
where $\eta$ is an auxiliary perturbation parameter, the influence function is
defined by
\begin{equation}
D(P_{\eta,x},B)
=
D(P,B)+\eta\,\IF_D(x;P)+o(\eta).
\label{eq:if-expansion-pop}
\end{equation}
Suppose a finite-sample estimator is asymptotically linear,
\begin{equation}
T_m
=
D(P,B)
+
\frac1m\sum_{i=1}^m \IF_D(X_i;P)
+
R_m,
\label{eq:asymptotic-linear}
\end{equation}
with
$\E[R_m\mid X_1=x]-\E[R_m]=o(m^{-1})$. 
Conditioning the above expression on one observation being fixed, $X_1=x$, and subtracting the unconditional expectation of $T_m$, we obtain
\begin{equation}
L_{T,m}(x)
=
\frac{1}{m}\,\mathrm{IF}_{D}(x;P)
+
o(m^{-1}).
\label{eq:localization-IF}
\end{equation}

This is the standard von Mises/influence-function expansion viewed through
the localization operation \cite{vonmises1947,hampel1974,vanderVaart1998}.
For the U-statistic functional generated by $h$, one has
$\IF_D(x;P)=r h_1(x)$, so Eq.~\eqref{eq:ustat-main} is its exact
finite-sample counterpart.

A useful boundary is also immediate.  If the first-order influence vanishes,
as it does for many squared discrepancies at their exact null, no nontrivial
single-event ranking is available at order $1/m$.  The relevant information
may then reside in the second or higher projections.  This is precisely what
happens for the unbiased MMD statistic at $P=B$.

\subsection{Maximum mean discrepancy and the witness function}
\label{sec:mmd}

MMD provides an exactly soluble example for our purposes. Its unbiased
estimator is a second-order U-statistic, so the finite-sample localization
result of Sec.~\ref{sec:theory} applies exactly. At the same time, MMD has a standard pointwise witness function that describes where the distributions $P$ and
$B$ differ~\cite{gretton2012,lloyd2015}. As shown below, these two objects are directly connected: the first Hoeffding projection of the MMD statistic is the centered witness function. We use the unnormalized witness
\begin{equation}
w_P(x)
=
\E_{X\sim P}k(x,X)-\E_{Y\sim B}k(x,Y).
\label{eq:unnormalized-witness}
\end{equation}
When $B$ is known, define
\[
\mu_B(x)=\E_{Y\sim B}k(x,Y),
\qquad
K_{BB}=\E_{Y,Y'\sim B}k(Y,Y'),
\]
and the centered pair kernel
\begin{equation}
q_B(x,y)
=
k(x,y)-\mu_B(x)-\mu_B(y)+K_{BB}.
\label{eq:mmd-U-kernel}
\end{equation}
Then
\begin{equation}
T_m^{\MMD}
=
\binom{m}{2}^{-1}\sum_{i<j}q_B(X_i,X_j)
\label{eq:mmd-u-pair}
\end{equation}
is unbiased for $\MMD^2(P,B)$.  Its first Hoeffding projection is
\[
h_1(x)=w_P(x)-\E_P[w_P(X)],
\]
so Eq.~\eqref{eq:ustat-main} gives the exact localization
\begin{equation}
\boxed{
L_{\MMD,m}(x)
=
\frac{2}{m}
\left[w_P(x)-\E_P w_P(X)\right].
}
\label{eq:mmd-localization}
\end{equation}
Equivalently,
\[
G_{\MMD,m}(x)
=
\MMD^2(P,B)
+
\frac{2}{m}
\left[w_P(x)-\E_P w_P(X)\right],
\]
so conditional resampling and the MMD witness induce exactly the same event ranking.

For the mixture model $P=(1-\epsilon)B+\epsilon S$,
\begin{equation}
\MMD^2(P,B)
=
\epsilon^2\MMD^2(S,B),
\qquad
w_P(x)
=
\epsilon\!\left[\E_S k(x,X)-\E_B k(x,Y)\right],
\label{eq:mmd-mixture-summary}
\end{equation}
and therefore
\begin{equation}
\boxed{
\E_S[L_{\MMD,m}(X)]-\E_B[L_{\MMD,m}(X)]
=
\frac{2\epsilon}{m}\MMD^2(S,B).
}
\label{eq:mmd-localization-class-separation}
\end{equation}
The distinction between global and local scaling is useful: the squared
global discrepancy is $O(\epsilon^2)$, while the witness and its class-mean
separation are $O(\epsilon)$.

At the exact null $P=B$, $w_B(x)=0$ for every fixed $x$, and hence
$L_{\MMD,m}(x)=0$ even for an observation deep in the background tail.  MMD
localization therefore measures support for a coherent distributional
departure, not isolated improbability.  The nonzero null fluctuations of the
global unbiased statistic reside in its degenerate second-order component.

\paragraph{Gaussian benchmark used in Experiment I.}
For the Gaussian kernel
$k_h(x,y)=\exp[-\|x-y\|^2/(2h^2)]$, the required kernel expectations between
Gaussian distributions are available in closed form \cite{rustamov2021}.
For
\[
B=\mathcal{N}(0,I_d), \qquad
S=\mathcal{N}(\mu,\sigma_s^2 I_d),
\]
where $I_d$ denotes the $d\times d$ identity matrix, we use
\begin{align}
a_B(x)
&=
\left(\frac{h^2}{h^2+1}\right)^{d/2}
\exp\!\left[-\frac{\|x\|^2}{2(h^2+1)}\right],
\\
a_S(x)
&=
\left(\frac{h^2}{h^2+\sigma_s^2}\right)^{d/2}
\exp\!\left[-\frac{\|x-\bm\mu\|^2}{2(h^2+\sigma_s^2)}\right],
\end{align}
and
\begin{align}
K_{BB}
&=\left(\frac{h^2}{h^2+2}\right)^{d/2},
\\
K_{SS}
&=\left(\frac{h^2}{h^2+2\sigma_s^2}\right)^{d/2},
\\
K_{SB}
&=\left(\frac{h^2}{h^2+1+\sigma_s^2}\right)^{d/2}
\exp\!\left[-\frac{\|\bm\mu\|^2}{2(h^2+1+\sigma_s^2)}\right].
\end{align}
Writing
$D_{SB}=K_{SS}+K_{BB}-2K_{SB}=\MMD^2(S,B)$, the contaminated-population
witness is
\[
w_P(x)=\epsilon[a_S(x)-a_B(x)],
\]
and the exact infinite-$R$ resampling score is
\begin{equation}
\boxed{
G_{\MMD,m}(x)
=
\epsilon^2D_{SB}
+
\frac{2}{m}
\left\{\epsilon[a_S(x)-a_B(x)]-\overline w_P\right\},
}
\label{eq:analytic-G-gaussian}
\end{equation}
where $\overline w_P=\E_P[w_P(X)]$.  These standard Gaussian identities are
used only as exact benchmarks for the numerical implementation.

\subsection{Efficient marginal and event-containing estimators}
\label{sec:marginal-contribution}

The population identities derived above show that conditional localization
and marginal-contribution constructions target the same event-level quantity,
or at least the same event ranking. This equivalence at the population level,
however, does not imply that their finite-resampling Monte Carlo estimators
have the same statistical efficiency. In particular, estimators based on
matched differences can cancel fluctuations due solely to the random context,
and can therefore have substantially smaller variance than a raw conditional
average.

  For a U-statistic of order $r$, separating terms that contain a
fixed $x$ from context-only terms gives
\begin{equation}
\Delta_m(x;C)
\equiv
U_m(x,C)-U_{m-1}(C)
=
\frac{r}{m}
\left[V^{(r-1)}_{m-1}(x;C)-U^{(r)}_{m-1}(C)\right],
\label{eq:U-context-difference}
\end{equation}
whose expectation is exactly $(r/m)h_1(x)=L_{T,m}(x)$.  The subtraction is
important because the two terms share the same random context.

For known-background MMD, use the centered pair kernel $q_B$ defined in
Eq.~\eqref{eq:mmd-U-kernel}.  The event-containing pair contribution in a
batch is
\begin{equation}
J_i
=
\frac{2}{m(m-1)}\sum_{j\neq i}q_B(x_i,x_j).
\label{eq:event-pair-contribution}
\end{equation}
If $T_{m-1}^{(-i)}$ is the statistic after removing $x_i$, then
\[
T_m-T_{m-1}^{(-i)}
=
J_i-\frac{2}{m}T_{m-1}^{(-i)}.
\]
Thus the generic marginal and pair forms carry the same localization
information, while the pair form removes an additional context-only term.

The variance reduction produced by matched subtraction can be seen directly
when the batch statistic $T_m$ is asymptotically linear, as in Eq.~\eqref{eq:asymptotic-linear}. In
that case its leading fluctuation can be written as an average of
single-observation contributions. For a fixed event $x$ and a random context
$C=(X_2,\ldots,X_m)$,
\[
T_m(x,C)
\simeq
D(P,B)
+
\frac{1}{m}
\left[
\mathrm{IF}_D(x;P)
+
\sum_{j=2}^{m}\mathrm{IF}_D(X_j;P)
\right].
\]
The random context therefore produces a variance of order $O(m^{-1})$ in the
raw conditional statistic.

In the matched difference
\[
\Delta_m(x;C)=T_m(x,C)-T_{m-1}(C),
\]
the same context appears in both terms. Its leading contribution is therefore
multiplied by
\[
\frac{1}{m}-\frac{1}{m-1}
=
-\frac{1}{m(m-1)},
\]
so that most of the context fluctuation cancels. The remaining leading context
variance is of order $O(m^{-3})$. In the generic first-order regime this gives
\begin{equation}
\frac{\operatorname{Var}[\Delta_m]}
     {\operatorname{Var}[T_m(x,C)]}
\simeq
\frac{1}{(m-1)^2}.
\label{eq:variance-reduction-factor}
\end{equation}
Near a degenerate null, however, the raw first-order term can be strongly
suppressed, so the observed ratio can show a pre-asymptotic $O(m^{-1})$
behavior; Experiment~I demonstrates this crossover explicitly.

For the pair estimator, conditioning on $x_i$ gives
\[
\Var(J_i\mid x_i)
\sim
\frac{4}{m^2(m-1)}
\Var[q_B(x_i,X)\mid x_i].
\]
In a global-batch implementation, where $N$ denotes the total number of
observations in the dataset, an event appears on average
\[
K \simeq \frac{Rm}{N}
\]
times among $R$ randomly drawn batches of size $m$.
After rescaling to witness normalization this yields
\begin{equation}
\Var(\widehat w_i)
\propto
\frac{1}{Rm^2},
\label{eq:pair-witness-variance-scaling}
\end{equation}
for the finite-resampling component of the error in the estimated
event-level witness.
To connect this variance to the correlation used below, write
\[
\hat w = w + \xi,
\]
where $\xi$ denotes the finite-resampling error in the estimated event-level
witness, and suppose that, to leading order, $\xi$ is uncorrelated with the
target witness $w$ across events. If
\[
\rho = \operatorname{Corr}(w,\hat w),
\]
then
\begin{equation}
q_\rho \equiv \rho^{-2}-1
=
\frac{\operatorname{Var}(\xi)}
     {\operatorname{Var}(w)}.
\label{eq:q-rho-definition}
\end{equation}
Therefore,
\begin{equation}
q_\rho \propto \frac{1}{Rm^2}.
\label{eq:q-rho-scaling}
\end{equation}
Experiment III tests this scaling directly.

\section{When does the ensemble contain additional class information?}
\label{sec:information}

The localization framework we discussed {\em supra} assigns a score to each event using a departure
estimated from an ensemble.  This does not by itself imply that the ensemble
contains class information unavailable from the event in isolation.  We now
separate these two issues.

\subsection{The fully specified IID case}
\label{sec:fully-specified-iid}

Consider a sample in which each observation $X_i$ has an unobserved class
label $Z_i\in\{B,S\}$, indicating whether the event is background or signal.
We assume
\[
\PP(Z_i=S)=\epsilon, \qquad \PP(Z_i=B)=1-\epsilon,
\]
and that, conditional on the class label,
\[
X_i \mid Z_i=B \sim b(x), \qquad
X_i \mid Z_i=S \sim s(x),
\]
where $b(x)$ and $s(x)$ denote the known background and signal densities,
respectively, and $\epsilon$ is the known signal fraction.
Assuming the events are IID, the marginal density of an observation is
\begin{equation}
p(x)=(1-\epsilon)b(x)+\epsilon s(x).
\label{eq:known-mixture-density}
\end{equation}
For one event,
\begin{equation}
\PP(Z_i=S\mid X_i=x)
=
\frac{\epsilon s(x)}
{(1-\epsilon)b(x)+\epsilon s(x)},
\label{eq:single-event-posterior}
\end{equation}
so the Bayes-optimal ranking is determined by $s(x)/b(x)$.

Let $X_{-i}$ denote all observations in the sample except $X_i$. We may ask
whether these additional observations change the posterior probability that
event $i$ is signal. Under the fully specified IID model they do not. Since
$X_{-i}$ is independent of $(Z_i,X_i)$,
\begin{equation}
P(Z_i=S\mid X_i,X_{-i})
=
P(Z_i=S\mid X_i).
\label{eq:no-ensemble-gain}
\end{equation}
The factors involving $X_{-i}$ cancel from Bayes' theorem because they are
the same whether $Z_i=S$ or $Z_i=B$. Equivalently,
\begin{equation}
I(Z_i;X_{-i}\mid X_i)=0.
\label{eq:conditional-MI-zero}
\end{equation}
Thus, once the complete IID mixture is specified, the rest of the sample
contains no additional information about the class of event $i$ beyond that
already contained in $X_i$ itself.

Note that the above statement concerns the label of one event; it does not say that additional IID observations provide no additional evidence for a global hypothesis.  For two fully specified global models, the ensemble likelihood
ratio factorizes into the product of the event-wise likelihood ratios, so
the evidence can grow with $N$ without introducing irreducibly new
multi-event information.  This distinction is closely related to the
single-event versus multi-event collider-classification analysis of
Ref.~\cite{nachman2021}.

Consequently, any genuinely contextual class information must arise because
at least one ingredient of the complete IID model is not known or because
the observations are not independent.  We focus on the simplest case:
an unknown alternative parameter shared by the signal population.

\subsection{A shared unknown alternative}
\label{sec:shared-alternative}

Suppose the signal belongs to a family $s(x\mid\theta)$ and the same unknown
$\theta$ applies to all signal events in a given experiment.  A hierarchical
description is
\begin{equation}
\theta\sim\pi(\theta),\qquad
Z_i\sim{\rm Bernoulli}(\epsilon),\qquad
X_i\mid Z_i,\theta\sim
\begin{cases}
b(x), & Z_i=B,\\
s(x\mid\theta), & Z_i=S.
\end{cases}
\label{eq:hierarchical-observation}
\end{equation}
Conditional on $\theta$, the events remain independent. If $\theta$ were known, the same argument as in Eq.~\eqref{eq:no-ensemble-gain} would therefore hold conditionally: once $X_i$ and $\theta$ are known, the remaining observations $X_{-i}$ do not change the posterior probability that event $i$ is signal, \begin{equation} \PP(Z_i=S\mid X_i,X_{-i},\theta) = \PP(Z_i=S\mid X_i,\theta). \label{eq:known-theta-no-context} \end{equation}
Equivalently, in terms of conditional mutual information,
\begin{equation}
I(Z_i;X_{-i}\mid X_i,\theta)=0,
\label{eq:known-theta-conditional-MI-zero}
\end{equation}
where $I(A;B\mid C)$ denotes the conditional mutual information
\cite{coverthomas2006}, which vanishes when knowledge of $B$ provides no additional information about $A$ once $C$ is known.

When $\theta$ is unknown, however, the rest of the dataset can constrain it.
To avoid using event $i$ to define the alternative against which it is
scored, let
\begin{equation}
\pi_{-i}(\theta)
\equiv
p(\theta\mid X_{-i})
\propto
\pi(\theta)
\prod_{j\neq i}
\left[(1-\epsilon)b(X_j)+\epsilon s(X_j\mid\theta)\right],
\label{eq:theta-loo-explicit}
\end{equation}
and define the posterior-predictive signal density
\begin{equation}
\widetilde s_{-i}(x)
=
\int s(x\mid\theta)\pi_{-i}(\theta)\,d\theta.
\label{eq:posterior-predictive-signal}
\end{equation}
Then
\begin{equation}
\PP(Z_i=S\mid X_i=x_i,X_{-i})
=
\frac{\epsilon\,\widetilde s_{-i}(x_i)}
{(1-\epsilon)b(x_i)+\epsilon\,\widetilde s_{-i}(x_i)}.
\label{eq:contextual-event-posterior}
\end{equation}
Without the ensemble, the corresponding density is the prior predictive
\begin{equation}
\overline s(x)
=
\int s(x\mid\theta)\pi(\theta)\,d\theta.
\label{eq:prior-predictive-signal}
\end{equation}

Thus the ensemble contains additional class information precisely when
learning from $X_{-i}$ changes the relevant predictive signal model on a set
of nonzero probability.  The natural measure is
\begin{equation}
I(Z_i;X_{-i}\mid X_i),
\label{eq:central-conditional-MI}
\end{equation}
which can be positive even though the events are conditionally independent
given $\theta$.  
Thus the ensemble contains additional class information precisely when
learning from $X_{-i}$ changes the relevant predictive signal model on a set
of nonzero probability. The other observations update the posterior
distribution of the shared parameter $\theta$ from $\pi(\theta)$ to
$\pi_{-i}(\theta)$. This in turn changes the posterior-predictive signal
density $\widetilde{s}_{-i}(x)$ against which $X_i$ is evaluated, and hence
can change the posterior probability that event $i$ is signal.

A useful interpretation follows directly from the chain rule and
Eq.~\eqref{eq:known-theta-no-context}:
\begin{equation}
\boxed{
I(Z_i;X_{-i}\mid X_i)
=
I(Z_i;\theta\mid X_i)
-
I(Z_i;\theta\mid X_i,X_{-i}).
}
\label{eq:oracle-gap-identity}
\end{equation}
The ensemble therefore recovers part, but never more than all, of the class
information that would be available if the hidden alternative were revealed
by an oracle.

The shared nature of $\theta$ is essential.  If instead each signal event
has an independent parameter $\theta_i\sim\pi(\theta)$, marginalization
produces IID signal draws from $\overline s(x)$; then the other observations
contain no information about the private $\theta_i$ of event $i$ and
Eq.~\eqref{eq:conditional-MI-zero} is recovered.

\subsection{A construction with zero single-event information}
\label{sec:zero-info}

The shared-parameter mechanism of Sec.~5.2 is quite general: the ensemble can
provide additional class information whenever the other observations constrain
a common signal parameter that is relevant to the interpretation of $X_i$.
We now construct a special Gaussian example in which this effect can be
isolated completely. The model is chosen so that signal and background have
exactly the same one-event marginal distribution, and hence an isolated event
contains no class information at all, while dependence induced by the shared
signal parameter remains visible at the ensemble level. Let
\begin{equation}
B={\cal N}(0,I_d),
\qquad
\bm{\mu}\sim
{\cal N}\!\left(0,(1-\sigma_s^2)I_d\right),
\qquad
S_{\bm{\mu}}
=
{\cal N}\!\left(\bm{\mu},\sigma_s^2I_d\right),
\label{eq:latent-gaussian-model}
\end{equation}
with $0<\sigma_s<1$, and define
\begin{equation}
\gamma\equiv1-\sigma_s^2.
\label{eq:gamma-definition}
\end{equation}
Within one pseudoexperiment all signal events share the same
$\bm{\mu}$.

Marginalizing over $\bm{\mu}$ gives
\begin{equation}
X_i\mid Z_i=S
\sim
{\cal N}\!\left(0,
[(1-\sigma_s^2)+\sigma_s^2]I_d\right)
=
{\cal N}(0,I_d)
=
X_i\mid Z_i=B.
\label{eq:signal-marginal-equals-background}
\end{equation}
Hence
\begin{equation}
I(Z_i;X_i)=0,
\qquad
\AUC_{\rm single}=0.5.
\label{eq:single-event-MI-zero}
\end{equation}
No transformation of an isolated observation can distinguish its class.

The information survives in the dependence between signal events.  For
$i\neq j$,
\begin{equation}
\boxed{
\mathrm{Cov}
\left(
X_i,X_j\mid Z_i=Z_j=S
\right)
=
\gamma I_d
=
(1-\sigma_s^2)I_d.
}
\label{eq:signal-cross-covariance}
\end{equation}
Equivalently, two signal observations are jointly Gaussian with marginal
covariances $I_d$ and cross-covariance $\gamma I_d$.  Standard Gaussian
conditioning therefore gives
\begin{equation}
X_j\mid X_i=x,Z_i=Z_j=S
\sim
{\cal N}\!\left(\gamma x,(1-\gamma^2)I_d\right).
\label{eq:signal-predictive-one-event}
\end{equation}

This already proves that one additional \emph{unlabelled} event can contain
class information about event $i$.  
If $Z_i=B$, then $X_i$ is independent of the latent signal location, and therefore
\begin{equation}
X_j\mid X_i=x,Z_i=B
\sim
\mathcal{N}(0,I_d).
\label{eq:companion-given-background}
\end{equation}
If $Z_i=S$, event $j$ is background with probability $1-\epsilon$ and signal with probability $\epsilon$. Therefore
\begin{equation}
X_j\mid X_i=x,Z_i=S
\sim
(1-\epsilon)\,\mathcal{N}(0,I_d)
+
\epsilon\,\mathcal{N}\!\left(
\gamma x,(1-\gamma^2)I_d
\right).
\label{eq:companion-given-signal}
\end{equation}
For $0<\epsilon<1$ and $0<\gamma<1$ these conditional distributions differ
on a set of nonzero probability.  Therefore
\begin{equation}
\boxed{
I(Z_i;X_i)=0,
\qquad
I(Z_i;X_{-i}\mid X_i)>0.
}
\label{eq:central-zero-positive-result}
\end{equation}
The second inequality follows already from the information carried by a
single companion $X_j$ and hence also holds for the full ensemble.

This construction also clarifies the connection with discrepancy
localization.  Within a particular pseudoexperiment, $\bm{\mu}$ is fixed
and
\begin{equation}
P_{\bm{\mu}}
=
(1-\epsilon)B+\epsilon S_{\bm{\mu}},
\end{equation}
so the MMD witness is
\begin{equation}
w_{\bm{\mu}}(x)
=
\epsilon
\left[
\E_{Y\sim S_{\bm{\mu}}}k(x,Y)
-
\E_{Y\sim B}k(x,Y)
\right],
\label{eq:mu-specific-witness}
\end{equation}
which is generally nonzero.  Across hypothetical experiments, however,
the prior-predictive signal distribution equals $B$, so
\begin{equation}
\E_{\bm{\mu}}[w_{\bm{\mu}}(x)]=0.
\label{eq:average-witness-zero}
\end{equation}
There is therefore no fixed, experiment-independent event score that can
solve the problem.  The relevant direction $P_{\bm{\mu}}-B$ must be learned
from the observed ensemble and then localized back onto its members.

As a control, if a new $\bm{\mu}_i$ is drawn independently for each signal
event from the same prior, all one-event marginals are unchanged but the
cross-event dependence disappears.  The observations are then IID and
$I(Z_i;X_{-i}\mid X_i)=0$.  Experiment~II compares these two constructions
directly.

\section{Estimation from finite samples}
\label{sec:algorithm}

The population localization $L_{T,m}(x)$ is defined with respect to the
unknown data-generating distribution $P$.  In practice it must be estimated
from a finite observed sample ${\cal D}_N=\{x_1,\ldots,x_N\}$.  This section
summarizes the estimators used below and the finite-sample issues that matter
for their interpretation.

\subsection{Historical resampling and conditional localization}
\label{sec:algorithm-localization}

The operational idea of Inverse Bagging \cite{vischia2017} is to generate
many small batches from a test sample, evaluate a goodness-of-fit statistic
against a background reference, and assign the batch information back to the
events that participated in it.  The original work considered both a binary
background-like/non-background-like decision and a continuous score.  The
latter may be written schematically as
\begin{equation}
\widehat G_i
=
\frac{1}{n_i}
\sum_{r:i\in B_r}T(B_r),
\label{eq:historical-average-score}
\end{equation}
where $n_i$ is the number of sampled batches containing event $i$.  In the
large-resampling limit this estimates the conditional average of the parent
statistic over contexts containing that event.

A cleaner finite-sample analogue of the population definition is obtained by
forcing $x_i$ to appear exactly once and drawing the remaining $m-1$
companions from ${\cal D}_N\setminus\{x_i\}$.  With $R$ independently sampled
contexts $C_{i,r}$,
\begin{equation}
\widehat G_i^{\rm force}
=
\frac1R\sum_{r=1}^{R}T_m(x_i,C_{i,r}).
\label{eq:forced-G-estimator}
\end{equation}
The companions may be sampled with or without replacement.  The former is
the direct bootstrap analogue of independent draws from $P$; the latter
avoids duplicate measured events and is natural for U-statistics built from
distinct observations.  Their difference vanishes as a finite-population
effect when $m/N$ is small.

The historical global-batch construction is computationally different.  A
single random batch contributes simultaneously to all events it contains.
For $R$ global batches of size $m$, a given event appears on average
\begin{equation}
K_i\simeq \frac{Rm}{N}
\label{eq:effective-context-small-m}
\end{equation}
times when $m\ll N$.  Thus $R$ denotes contexts \emph{per event} in the
forced-inclusion construction but total global batches in the historical
construction; the distinction is important when comparing Monte Carlo
scaling.  Bootstrap multiplicities also mean that an event can occur more
than once in a global batch.  We avoid this ambiguity in the controlled
forced-inclusion studies and retain the historical bootstrap construction in
Experiment~III, where reproducing its computational behaviour is part of the
comparison.

For fixed data, finite $R$ produces only Monte Carlo error and the usual
standard error decreases as $R^{-1/2}$ (or as $K_i^{-1/2}$ for a global-batch
score).  Increasing $R$ cannot remove finite-$N$ fluctuations of the
empirical distribution.  We therefore distinguish convergence of the
resampling estimator on a fixed dataset from convergence of the empirical
localization itself to its population limit.

\subsection{Direct estimation, self-influence and scalable approximations}
\label{sec:direct-estimation}

For MMD the population localization is already known analytically from
Sec.~\ref{sec:mmd}: event ranking is exactly the ranking of the witness
$w_P(x)$.  The natural empirical counterpart is
\begin{equation}
\widehat w_N(x)
=
\frac1N\sum_{j=1}^{N}k(x,x_j)-\mu_B(x),
\qquad
\mu_B(x)=\E_{Y\sim B}k(x,Y).
\label{eq:empirical-witness}
\end{equation}
For MMD, the event ordering corresponding to the infinite-resampling limit can be obtained directly from the empirical witness, so explicit resampling is not needed for the ranking itself. We nevertheless perform the resampling because MMD provides a controlled case in which the generic localization procedure can be compared with a directly calculable event-level target.

When $x_i$ is scored with Eq.~\eqref{eq:empirical-witness}, 
it also appears among the observations
used to construct the empirical witness itself, through the term
$k(x_i,x_i)$. To remove this direct self-contribution, we therefore use the leave-one-out (LOO) witness
\begin{equation}
\widehat w_{-i}(x_i)
=
\frac{1}{N-1}\sum_{j\neq i}k(x_i,x_j)-\mu_B(x_i).
\label{eq:loo-witness}
\end{equation}
For a fixed kernel and background model,
\begin{equation}
\E\!\left[\widehat w_{-i}(x_i)\mid X_i=x\right]=w_P(x),
\label{eq:loo-unbiased}
\end{equation}
whereas the self-inclusive score differs by an event-dependent term of order
$N^{-1}$:
\begin{equation}
\widehat w_N(x_i)-\widehat w_{-i}(x_i)
=
\frac1N
\left[
 k(x_i,x_i)
-
\frac{1}{N-1}\sum_{j\neq i}k(x_i,x_j)
\right].
\label{eq:self-difference}
\end{equation}
This distinction matters under the exact null.  For a Gaussian kernel,
$k(x,x)=1$, so a self-inclusive empirical witness can assign a small positive
score to an isolated tail event even though the population witness is zero;
the LOO construction removes that direct self-use.  Experiment~I uses this
property as a sharp validation test.

Leave-one-out removes the direct contribution of $x_i$ to its own empirical
witness, and is sufficient when all other ingredients of the scoring procedure
--- such as the kernel, its bandwidth, the event representation, and the
background model --- have been fixed independently of $x_i$. If some of these
ingredients are instead estimated from the same dataset, simply removing
$x_i$ from the final kernel sum is not sufficient, because the event may
already have influenced the scoring rule indirectly. In that case one should
use cross-fitting or an explicit sample split: the scoring rule applied to an
event is constructed using data from which that event, or its entire fold, has
been excluded~\cite{chernozhukov2018}. This controls such self-use and the
associated overfitting. It does not, however, correct for a subsequent
selection of unusually large scores or other data-dependent choices; that
separate post-selection issue is discussed in Sec.~\ref{sec:global-local}.

Exact evaluation of all LOO kernel sums costs $O(N^2)$.  Experiment~III
therefore uses random Fourier features (RFF) \cite{rahimi2007}.  For a
shift-invariant kernel one constructs $z(x)\in\mathbb R^D$ such that
\begin{equation}
k(x,y)\simeq z(x)^\top z(y).
\label{eq:rff-kernel}
\end{equation}
With
\[
\bar z=\frac1N\sum_{j=1}^{N}z(x_j),
\qquad
\bar z_B=\E_{Y\sim B}z(Y),
\]
the empirical witness becomes
\begin{equation}
\widehat w_N(x)
\simeq
z(x)^\top(\bar z-\bar z_B),
\label{eq:rff-witness}
\end{equation}
and the LOO version is
\begin{equation}
\widehat w_{-i}^{\rm RFF}(x_i)
=
z(x_i)^\top
\left[
\frac{N\bar z-z(x_i)}{N-1}-\bar z_B
\right].
\label{eq:rff-loo}
\end{equation}
After the feature map has been evaluated, all scores can be obtained in
$O(ND)$ time.  The RFF dimension $D$ then controls a transparent
computational--approximation trade-off.

\subsection{Raw, marginal, and event-containing estimators}
\label{sec:practical-estimators}

The population identities of Secs.~\ref{sec:marginal-principle} and
\ref{sec:marginal-contribution} suggest three generic finite-resampling
estimators.  For event $x_i$ and matched random contexts $C_r$,
\begin{align}
\widehat G_i^{\rm raw}
&=
\frac1R\sum_{r=1}^R T_m(x_i,C_r),
\label{eq:three-G}
\\
\widehat A_i^{\rm add}
&=
\frac1R\sum_{r=1}^R
\left[T_m(x_i,C_r)-T_{m-1}(C_r)\right],
\label{eq:three-add}
\\
\widehat A_i^{\rm rep}
&=
\frac1R\sum_{r=1}^R
\left[T_m(x_i,C_r)-T_m(Y_r,C_r)\right].
\label{eq:three-rep}
\end{align}
The raw score estimates the conditional statistic directly.  The marginal
scores subtract a matched baseline and therefore cancel fluctuations that
belong only to the random context.  For pairwise U-statistics such as MMD,
the event-containing interaction
\begin{equation}
J_i
=
\frac{2}{m(m-1)}\sum_{j\neq i}q_B(x_i,x_j)
\end{equation}
provides a still more direct specialization by accumulating only terms that
actually contain the event.

These estimators have the same population localization information but very
different Monte Carlo efficiency.  Section~\ref{sec:marginal-contribution}
showed that matched subtraction can reduce the generic context variance from
$O(m^{-1})$ to $O(m^{-3})$ and that, in the global-batch MMD implementation,
the witness-normalized pair estimator obeys
\[
\Var(\widehat w_i)\propto (Rm^2)^{-1}.
\]
The raw, marginal and event-containing estimators are compared directly in
Experiments~I--III.  For MMD, the exact LOO witness provides the finite-data
convergence target against which their stochastic errors can be measured.

\subsection{Global discovery and event localization}
\label{sec:global-local}

A global test of
\[
H_0:P=B
\]
and the localization of the observations responsible for a detected
departure are distinct statistical tasks.  A large localization score is not
by itself evidence against $H_0$, and it is not a posterior probability that
the corresponding event is signal.  In this work the localized scores are
used primarily as descriptive rankings of how strongly individual events
align with the departure defined by the chosen discrepancy.

When a discovery significance is required to put forth a scientific claim, the global statistic and every data-dependent choice entering it must be calibrated under the null.  For
the nonstandard statistics considered here, background-only
pseudoexperiments provide a direct route: the same bandwidth scans,
selection rules or other optimizations performed on the observed data must be repeated in each pseudoexperiment.  This automatically incorporates the
relevant look-elsewhere effects \cite{grossvitells2010}.  Standard
likelihood-based asymptotics may of course be used when their conditions are
satisfied \cite{cowan2011}.

If localized events are subsequently used for a second formal inferential
stage, the adaptive selection must also be accounted for.  Two transparent
options are an independent validation sample or full null calibration of the
complete discovery--localization--selection pipeline; selective-inference
methods provide a more formal alternative \cite{fithian2014,lee2016}.
Cross-fitting prevents an event from helping to construct its own score, but
it does not by itself correct the fact that one may subsequently select the
largest score among many events.

Finally, a realistic collider search must account for uncertainty in the
background model in both the global discrepancy statistic and its event-level
localization. If the assumed background distribution differs systematically
from the true one, this mismatch can itself produce a nonzero, spatially
coherent witness and may therefore be mistaken for a genuine signal-like
departure.
The controlled studies below deliberately separate this issue from the
localization mechanism: $B$ is treated as known, the event count $N$ is
fixed, and event scores are evaluated against known labels without assigning
individual discovery $p$-values.  The numerical experiments therefore test
the question central to this work: once a distributional departure exists,
how efficiently can its event-level localization be recovered?


\section{Numerical studies}
\label{sec:toys}

The three experiments we provide in this section are designed to test different claims of the framework.
Experiment~I validates the population localization identities and the variance
reduction of matched estimators in a fully controlled model.  Experiment~II then 
asks the distinct information-theoretic question of whether an ensemble can
contain class information absent from an isolated event.  Experiment~III finally
tests the estimator scaling on a realistic collider benchmark. It is important to note here that the focus of these numerical studies is on testing the specific theoretical statements made {\em supra} rather
than to optimize anomaly-detection performance.

\subsection{Experiment I: controlled validation with known signal}
\label{sec:exp-known-signal}

The first experiment uses a fixed and known alternative, so no genuinely
collective class information is involved.  Its purpose is to validate the
conditional-localization formalism, separate population from finite-sample and
Monte Carlo effects, and test the variance cancellation of the marginal and
event-containing estimators.

We take
\[
B=\mathcal N(0,I_d),
\qquad
S=\mathcal N(\bm\mu_0,\sigma_s^2 I_d),
\]
with
\[
d=2,\qquad N=5000,\qquad \epsilon=0.02,\qquad
\bm\mu_0=(1.5,0),\qquad \sigma_s=0.5,
\]
and fix $N_S=100$ signal and $N_B=4900$ background events in each
pseudoexperiment.  Conditioning on the class counts suppresses composition
fluctuations in this estimator-validation study; the population identities
tested below refer to the same mixture parameters.  The Gaussian MMD kernel
has bandwidth $h=1$.  For this
model the conditional score is known analytically from
Eq.~\eqref{eq:analytic-G-gaussian}:
\begin{equation}
G_m(x)
=
\MMD^2(P,B)
+
\frac{2}{m}
\left[w_P(x)-\E_Pw_P(X)\right].
\label{eq:exp1-G}
\end{equation}
Thus event ranking is exactly the population-witness ranking for every fixed
$m$.  The parameters give
\[
\MMD^2(S,B)=0.46086,
\]
and, for $m=20$,
\begin{equation}
\Delta G
\equiv
\E_S[G_m]-\E_B[G_m]
=
\frac{2\epsilon}{m}\MMD^2(S,B)
=
9.2172\times10^{-4}.
\label{eq:exp1-deltag}
\end{equation}

\paragraph{Population identity and explicit resampling.}
As a first check, 500 independent pseudoexperiments give
\begin{equation}
\left\langle\Delta G\right\rangle
=
(9.2287\pm0.0125)\times10^{-4},
\end{equation}
corresponding to
\begin{equation}
\left\langle
\frac{\Delta G_{\rm toy}}{\Delta G_{\rm theory}}
\right\rangle
=
1.00124\pm0.00136,
\end{equation}
with pseudoexperiment-to-pseudoexperiment RMS $0.0304$.  The analytic
class-separation identity is therefore reproduced directly by the simulated
ensembles.

\begin{figure}[t]
    \centering
    \includegraphics[width=0.82\textwidth]{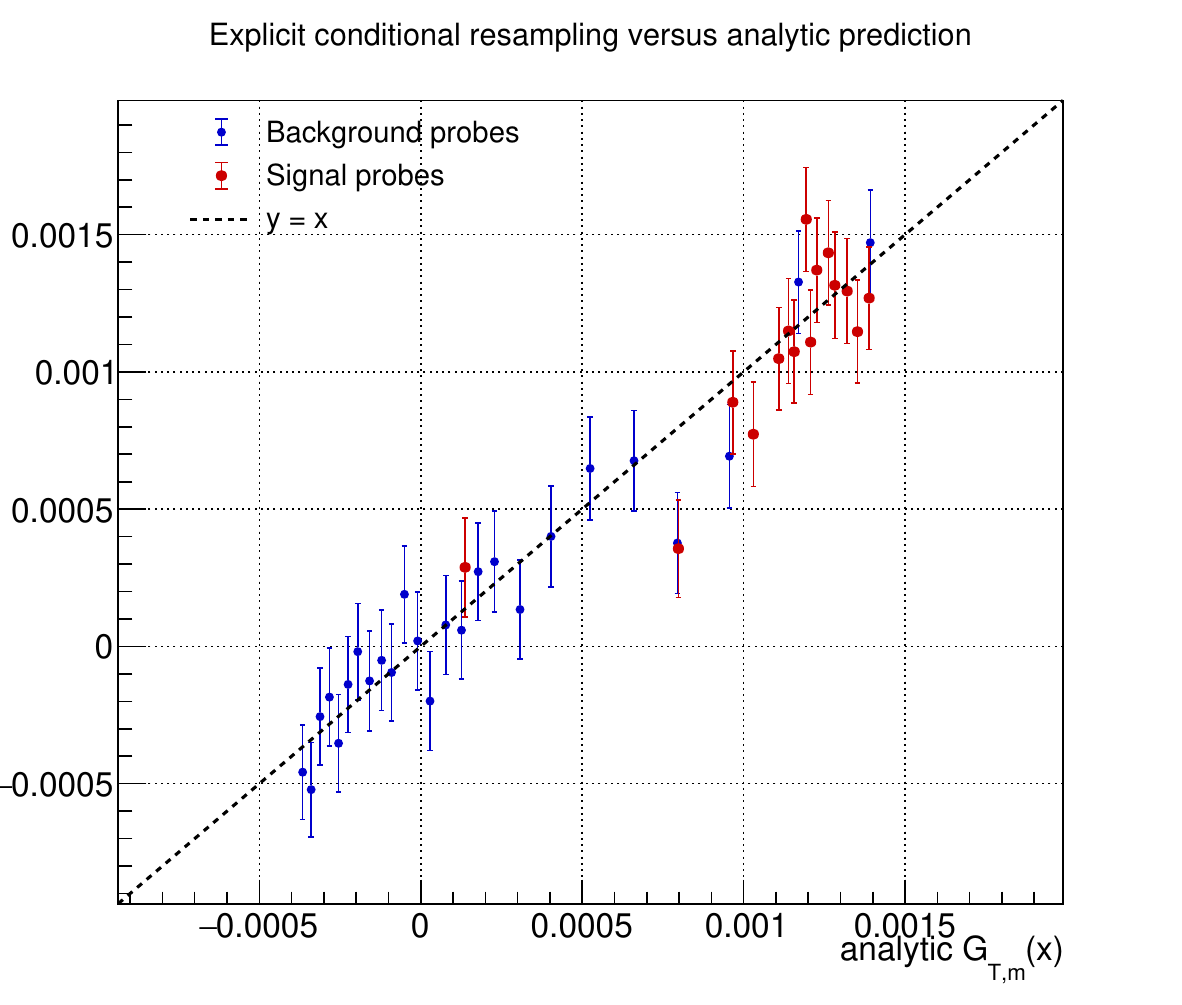}
    \caption{
    Direct validation of conditional MMD localization by explicit
    resampling.  Event scores obtained by forcing individual events
    into repeatedly resampled batches are compared with the analytic
    conditional expectation of Eq.~\eqref{eq:exp1-G}.
    Signal and background events follow the same predicted relation,
    showing that explicit resampling recovers the conditional projection
    of the batch statistic rather than defining an independent anomaly score.
    }
    \label{fig:exp1-resampled-analytic}
\end{figure}

We next force selected events into repeated batches and redraw the remaining
$m-1$ companions from the population.  Figure~\ref{fig:exp1-resampled-analytic}
compares the resulting Monte Carlo score with the exact conditional expectation
in Eq.~\eqref{eq:exp1-G}.
The RMS residual decreases with the expected $R^{-1/2}$ Monte Carlo scaling
over more than two orders of magnitude in the number of contexts $R$.

A complementary null test probes the distinction between isolated
improbability and membership in a distributional discrepancy.  We fix the
extreme observation
\[
x^\star=(5,0)
\]
and draw every companion from $B$.  Although $x^\star$ lies deep in the
background tail, first-order MMD localization predicts zero.  With $10^4$
contexts we obtain
\begin{equation}
\widehat G(x^\star)
=
(0.007\pm0.173)\times10^{-3},
\end{equation}
consistent with zero.  The test confirms numerically that MMD localization is
sensitive to support for a coherent departure, not simply to small background
density.

\begin{figure}[t]
    \centering
    \includegraphics[width=0.75\textwidth]{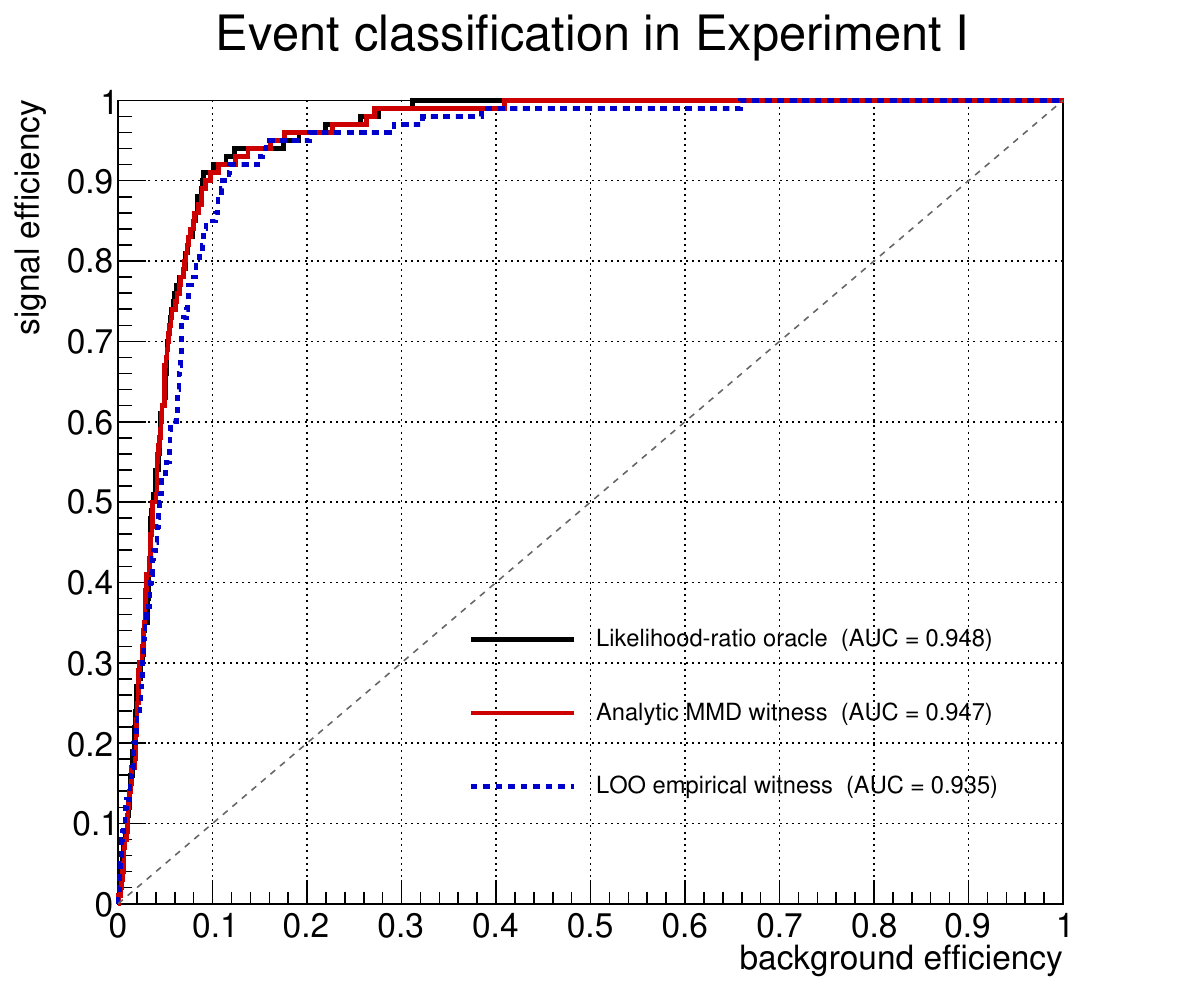}
\caption{
Event-level ROC curves for one representative pseudoexperiment in the
controlled fixed-alternative study.  Ensemble-averaged results are
reported in the text.
}
\label{fig:exp1-roc}
\end{figure}

\paragraph{Event ranking and finite-sample witness estimation.}
We compare the exact likelihood-ratio oracle, the population MMD witness, and
the empirical LOO witness of Eq.~\eqref{eq:loo-witness}.  For one representative
pseudoexperiment, Fig.~\ref{fig:exp1-roc} gives AUC values $0.948$, $0.947$,
and $0.935$, respectively.  Across 500 pseudoexperiments the two population
scores give
\begin{align}
\left\langle\AUC_{\rm LR}\right\rangle
&=0.93685\pm0.00032,\\
\left\langle\AUC_{\rm MMD}\right\rangle
&=0.93494\pm0.00035,
\end{align}
with paired difference
\begin{equation}
\left\langle\AUC_{\rm LR}-\AUC_{\rm MMD}\right\rangle
=0.00191\pm0.00008.
\end{equation}
Thus for this Gaussian alternative the population MMD witness is close to the
Bayes-optimal event ordering.

The empirical witness shows a larger finite-sample degradation.  In 100
pseudoexperiments for which the exact LOO kernel sums were evaluated,
\begin{equation}
\left\langle\AUC_{\rm LOO}\right\rangle
=0.9023\pm0.0043,
\end{equation}
with
\begin{equation}
\left\langle\AUC_{\rm MMD}-\AUC_{\rm LOO}\right\rangle
=0.0335\pm0.0043.
\end{equation}
The loss varies substantially among pseudoexperiments because the empirical
witness error is a correlated random field rather than independent
point-by-point noise.  In an extreme realization the empirical AUC fell to
$0.691$; the residual field showed a negative deformation around the signal
region and a positive deformation on the opposite side of the background.
This explains how a pointwise unbiased localizer can nevertheless suffer a
coherent finite-sample distortion of its event ranking.

\begin{figure}[t]
    \centering
    \includegraphics[width=\textwidth]{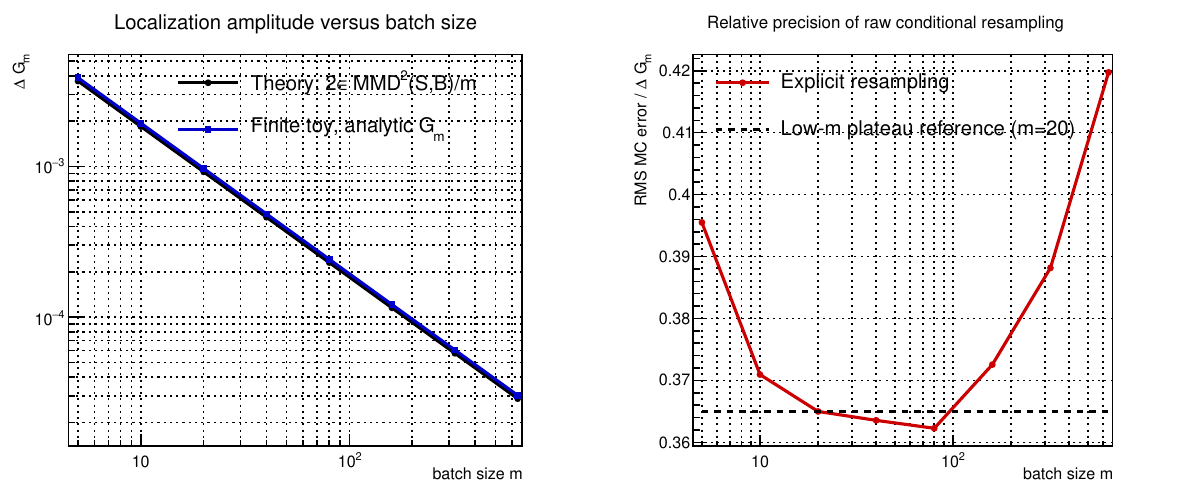}
    \caption{
    Dependence of the MMD localization score and raw-resampling uncertainty on
    batch size $m$, for $R=3000$ forced-inclusion contexts per probe event.
    \textit{Left:} signal--background localization separation; both the
    analytic prediction and finite-pseudoexperiment result follow the exact
    $1/m$ scaling.  \textit{Right:} RMS Monte Carlo uncertainty relative to
    $\Delta G_m$.  The relative error initially falls, reaches a minimum near
    $m\simeq80$, and then rises.  This U-shaped behavior marks the crossover
    from a near-degenerate second-order regime toward the generic first-order
    Hoeffding regime.
    }
    \label{fig:exp1-mscan}
\end{figure}

\paragraph{Batch size and the near-degenerate regime.}
Equation~\eqref{eq:exp1-G} predicts $\Delta G_m\propto1/m$, which we test over
\[
m=5,10,20,40,80,160,320,640.
\]
The left panel of Fig.~\ref{fig:exp1-mscan} confirms the exact $1/m$ scaling.
Because $m$ changes only the positive normalization of the population
localizer, the population event ordering is unchanged.

The right panel tests a different issue: how the stochastic error changes as
the parent MMD U-statistic moves away from its near-degenerate regime.  At
fixed $R=3000$, the relative RMS error decreases from $0.396$ at $m=5$ to a
minimum of approximately $0.362$ at $m=80$, then rises to $0.373$, $0.388$,
and $0.420$ for $m=160$, $320$, and $640$.  The behavior is consistent with
the variance structure discussed in Sec.~\ref{sec:marginal-contribution}: at
small and intermediate $m$ the suppressed first projection leaves
second-order fluctuations important, whereas at larger $m$ the first-order
term begins to control the relative precision.  The fully first-order
asymptotic regime is not yet reached at $m=640$.

\paragraph{Marginal and event-containing estimation.}
We finally test the variance cancellation central to the proposed estimator.
For every probe event and every freshly generated context we evaluate the raw
conditional statistic, the matched addition contribution of
Eq.~\eqref{eq:three-add}, and the event-containing pair term of
Eq.~\eqref{eq:event-pair-contribution}, using the same random companions.  All
three have the same signal--background population localization information;
only their context noise differs.

Figure~4 uses $R=1000$ forced-inclusion contexts per probe event and
$m=10,20,40,80,160,320$. Over this range, the relative uncertainty of the raw estimator is approximately independent of $m$, whereas the marginal and event-containing estimators decrease approximately as $m^{-1/2}$. The latter behavior is the expected relative scaling when the efficient context variance is $O(m^{-3})$ and the localization signal is $O(m^{-1})$.


\begin{figure}[t]
    \centering
    \includegraphics[width=\textwidth]{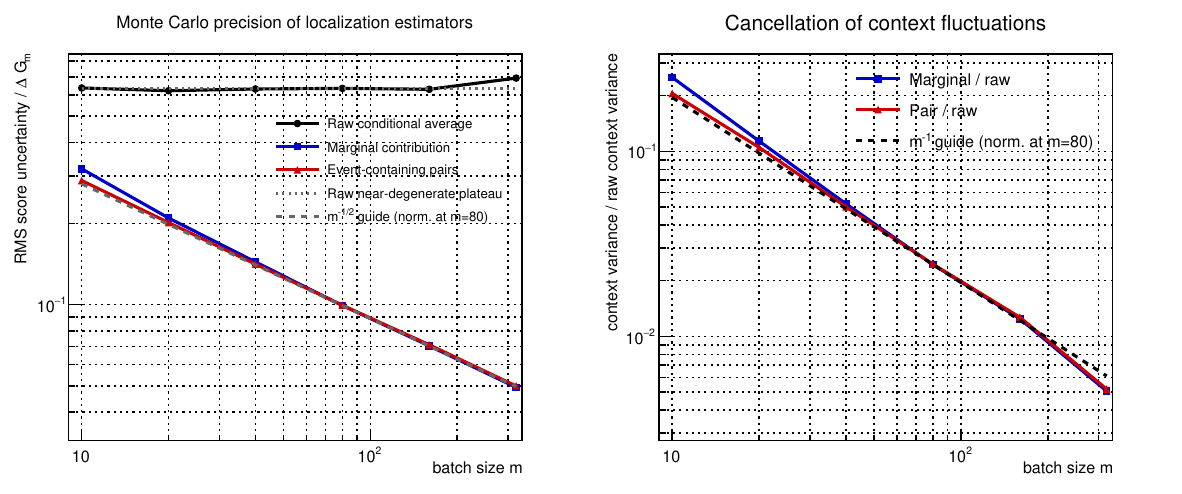}
    \caption{
    Monte Carlo efficiency of marginal and event-containing MMD localization
    in the controlled Gaussian experiment, using $R=1000$ forced-inclusion
    contexts per probe event.  \textit{Left:} RMS score uncertainty divided by
    the common class separation $\Delta G_m$.  Raw resampling remains on the
    near-degenerate plateau, whereas marginal and pair estimators closely
    follow the $m^{-1/2}$ reference.  \textit{Right:} variance ratios relative
    to raw averaging.  They fall approximately as $m^{-1}$ because the raw MMD
    statistic is still in its $O(m^{-2})$ near-degenerate regime, while the
    efficient estimators scale as $O(m^{-3})$.
    }
    \label{fig:exp1-marginal-pair}
\end{figure}

Over the scanned range, both variance ratios decrease approximately as
$m^{-1}$. This does not contradict the generic $m^{-2}$ ratio of
Eq.~\eqref{eq:variance-reduction-factor}, which assumes that the raw statistic has
already reached its nondegenerate $O(m^{-1})$ variance regime. Here the raw
MMD statistic remains predominantly second order, with variance approximately
$O(m^{-2})$, while the matched estimators have variance approximately
$O(m^{-3})$; their ratio therefore scales approximately as $m^{-1}$.
The marginal and pair estimators are already nearly indistinguishable by
$m\simeq40$ and have essentially identical precision for $m\geq80$.


Experiment~I therefore validates both the population target and its efficient
estimation: explicit resampling reproduces the conditional projection, the MMD
witness provides an effective event ranking for a fixed alternative, and
matched marginal/event-containing estimators remove most of the irrelevant
context fluctuation.  We next isolate the separate question of genuinely
collective class information.

\FloatBarrier
\subsection{Experiment II: zero isolated-event information}
\label{sec:exp-zero-info}

Experiment~II uses the shared-latent Gaussian construction of
Sec.~\ref{sec:zero-info}.  By construction, signal and background have
identical one-event marginals and every isolated-event classifier has
$\AUC=0.5$; any systematic discrimination must therefore come from information
shared across observations.  We use
\[
d=2,\qquad \epsilon=0.02,\qquad \sigma_s=0.5,\qquad h=1,
\]
with $N_S=\epsilon N$ fixed in each pseudoexperiment.  This conditions on the
total signal count rather than drawing the labels independently as in
Sec.~\ref{sec:fully-specified-iid}, and is used to suppress composition
fluctuations.  
Fixing the total number of signal events introduces dependence among the unobserved labels: knowing the labels of the other events would constrain the label of event $i$. This does not, however, generate contextual information in the observed features of the independent-$\mu_i$ control.
After marginalizing over each private $\mu_i$, both signal and background events have distribution $B$. Consequently, the feature vectors of the remaining events contain no information about their latent labels, and the fixed-count constraint cannot be exploited to classify event $i$.

For each pseudoexperiment we compare three scores: the likelihood-ratio oracle that knows the realized shared displacement $\bm\mu$, the population MMD
witness for that same $\bm\mu$, and the empirical LOO witness, which must infer
the departure from the remaining observations.  The critical control redraws
an independent $\bm\mu_i$ for every signal event.  This leaves every one-event
marginal unchanged while removing the shared cross-event structure, so the two
ensembles differ only in whether contextual class information exists.

\paragraph{Shared versus independent latent structure.}
For $N=5000$, 500 pseudoexperiments give the known-$\bm\mu$ benchmarks
\begin{align}
\left\langle\AUC_{\rm LR,true\text{-}\mu}\right\rangle
&=0.8884\pm0.0024,\\
\left\langle\AUC_{\rm MMD,true\text{-}\mu}\right\rangle
&=0.8846\pm0.0025.
\end{align}
The exact empirical LOO calculation, performed for 50 pseudoexperiments, gives
\begin{equation}
\left\langle\AUC_{\rm LOO,shared}\right\rangle
=0.8036\pm0.0209,
\end{equation}
whereas the independent-latent control gives
\begin{equation}
\left\langle\AUC_{\rm LOO,independent}\right\rangle
=0.5006\pm0.0048.
\end{equation}
Their difference is
\begin{equation}
\left\langle
\AUC_{\rm LOO,shared}-\AUC_{\rm LOO,independent}
\right\rangle
=0.3029\pm0.0224.
\end{equation}
The gain therefore does not arise merely from estimating an event score using
many observations: it appears only when those observations share information
about the alternative.

Performance varies across pseudoexperiments because $\bm\mu$ itself is random.
Small $\|\bm\mu\|$ produces an intrinsically weak departure and drives both
oracle and empirical AUCs toward $0.5$; for larger displacements the empirical
witness increasingly approaches the known-$\bm\mu$ benchmark.

\paragraph{Accumulation of contextual information with ensemble size.}
To test that interpretation directly, we repeat the construction at
\[
N=500,1000,2000,5000,10000,20000,
\]
using the same realized $\bm\mu$ across $N$ within each paired
pseudoexperiment.  The true-$\bm\mu$ likelihood-ratio and MMD benchmarks remain
approximately constant at $\AUC\simeq0.88$, and the independent-$\bm\mu_i$
control remains consistent with $0.5$.  The shared-context empirical witness,
by contrast, evolves as
\begin{equation}
\begin{array}{c|cccccc}
N&500&1000&2000&5000&10000&20000\\ \hline
\AUC_{\rm LOO,shared}&0.680&0.675&0.742&0.819&0.823&0.857
\end{array}
\end{equation}
and progressively approaches the known-alternative performance.

\begin{figure}[t]
    \centering
    \includegraphics[width=\textwidth]{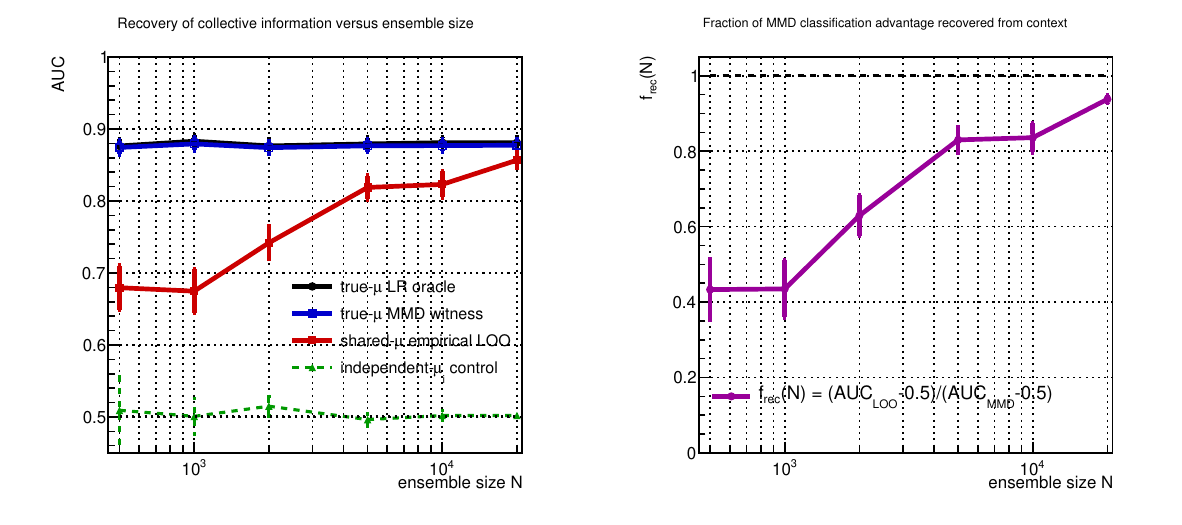}
    \caption{
    Recovery of collective classification information with ensemble size in
    the shared-latent experiment.  \textit{Left:} mean AUC for the true-$\bm\mu$
    likelihood-ratio oracle, true-$\bm\mu$ MMD witness, empirical LOO witness,
    and independent-$\bm\mu_i$ control.  Only the shared-context empirical
    witness improves systematically with $N$.  \textit{Right:} fraction
    $f_{\rm rec}$ of the MMD classification advantage above chance recovered by
    the empirical contextual witness.  Error bars are standard errors across
    pseudoexperiments.
    }
    \label{fig:exp2-Nscan}
\end{figure}

The paired gap to the true-$\bm\mu$ MMD witness decreases from
\begin{equation}
\Delta_{\rm MMD}(500)=0.1948\pm0.0259
\end{equation}
to
\begin{equation}
\Delta_{\rm MMD}(20000)=0.0208\pm0.0032,
\end{equation}
where
$\Delta_{\rm MMD}(N)=\AUC_{\rm MMD,true\text{-}\mu}(N)-
\AUC_{\rm LOO,shared}(N)$.  Defining the recovered fraction
\begin{equation}
f_{\rm rec}(N)
=
\frac{\AUC_{\rm LOO,shared}(N)-0.5}
{\AUC_{\rm MMD,true\text{-}\mu}(N)-0.5},
\end{equation}
we obtain
\begin{equation}
f_{\rm rec}(20000)=0.938\pm0.010.
\end{equation}
Thus the empirical contextual witness recovers about $94\%$ of the MMD
classification advantage available when the hidden alternative is known.  The
result is the numerical counterpart of Sec.~\ref{sec:zero-info}: increasing
$N$ helps only when the additional observations constrain a structure shared
with the event being classified.

\paragraph{Efficient recovery of the contextual score.}
The preceding results establish that contextual information exists.  We next
ask whether the variance-reduced estimators recover the same finite-sample
localizer from random global batches.  We use a representative $N=5000$
shared-$\bm\mu$ pseudoexperiment with $\|\bm\mu\|=0.7723$, batch size $m=100$,
and sample without replacement.  We vary
\[
R=10^3,3\times10^3,10^4,3\times10^4,10^5
\]
global batches; at the largest budget each event appears in about $2000$
batches on average.

For this pseudoexperiment the true-$\bm\mu$ MMD witness has
\[
\AUC_{\rm MMD,true\text{-}\mu}=0.8448,
\]
while the exact empirical LOO witness gives
\[
\AUC_{\rm LOO,shared}=0.7584.
\]
The exact finite-dataset pair target has
\begin{equation}
\AUC_{\rm pair,target}=0.7580,
\qquad
\rho(\text{pair target},\widehat w_{-i})=0.99997,
\end{equation}
showing that its ordering is effectively identical to the ordinary LOO
witness.  At $R=10^5$ the Monte Carlo estimators give
\begin{align}
\AUC_{\rm marginal}&=0.7604,
&\rho(\widehat A,\widehat w_{-i})&=0.9711,\\
\AUC_{\rm pair}&=0.7593,
&\rho(\widehat J,\widehat w_{-i})&=0.9718.
\end{align}
The small excursions above the exact target are compatible with residual Monte
Carlo error and are not interpreted as improved discrimination.

The independent-$\bm\mu_i$ pseudoexperiment supplies the essential control on
this convergence test.  Its particular finite realization has
$\AUC_{\rm LOO}=0.51769$ and pair-target $\AUC=0.51777$, with correlation
$0.99999$.  Such a single realization need not have exactly chance AUC because
accidental empirical fluctuations can correlate weakly with the labels; the
ensemble result $0.5006\pm0.0048$ above is what establishes the absence of
systematic class information.  At $R=10^5$ the marginal and pair estimators
converge to this accidental finite-sample target:
\begin{align}
\AUC_{\rm marginal}&=0.51782,
&\rho(\widehat A,\widehat w_{-i})&=0.99493,\\
\AUC_{\rm pair}&=0.51775,
&\rho(\widehat J,\widehat w_{-i})&=0.99500.
\end{align}
Repeated contextual averaging therefore reconstructs the empirical departure;
it does not manufacture class information.  Systematic alignment with the
labels appears only in the shared-latent model.

\begin{figure}[t]
    \centering
    \includegraphics[width=\textwidth]{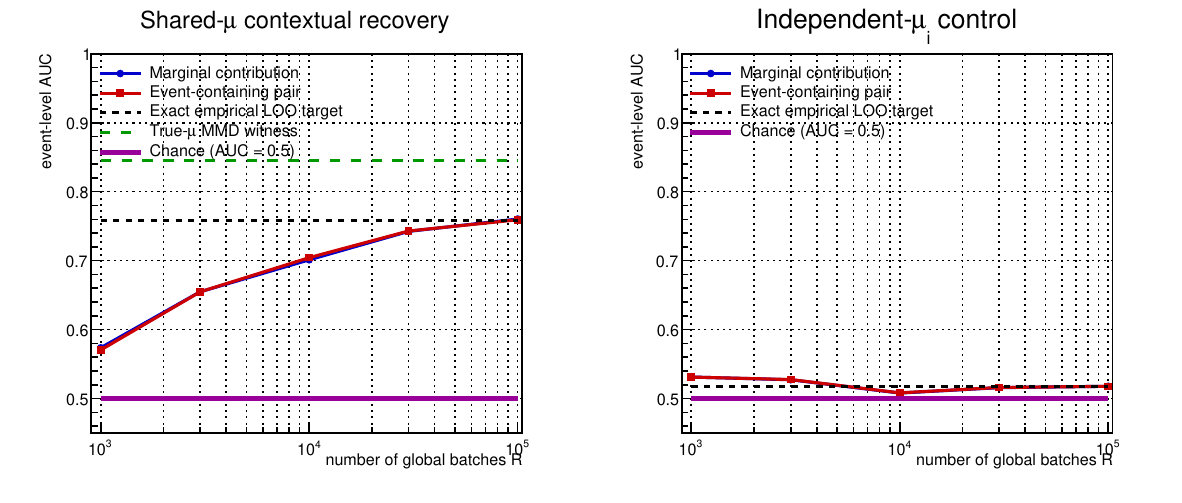}
    \caption{
    Recovery of the empirical contextual score by marginal and event-containing
    estimators for $N=5000$ and $m=100$.  \textit{Left:} shared-$\bm\mu$
    construction.  Both Monte Carlo estimators approach the empirical LOO
    target ($\AUC=0.7584$); the true-$\bm\mu$ MMD witness ($\AUC=0.8448$) shows
    the remaining finite-ensemble loss.  \textit{Right:} independent-$\bm\mu_i$
    control.  The estimators converge to the accidental finite-sample LOO value
    ($\AUC\simeq0.5177$), while the ensemble-average control remains consistent
    with chance.  The marginal and pair estimators are nearly indistinguishable
    in both cases.
    }
    \label{fig:exp2-marginal-recovery}
\end{figure}

\FloatBarrier
\subsection{Experiment III: LHC Olympics benchmark and estimator scaling}
\label{sec:exp-lhco}

The third experiment moves from controlled Gaussian models to a realistic
collider benchmark.  We use the LHC Olympics R\&D dataset
\cite{lhcolympics2021} and the same signal-region definition and four auxiliary
variables used in the CATHODE benchmark \cite{hallin2022}:
\[
3.3<m_{JJ}<3.7~{\rm TeV},
\qquad
x=(m_{J,\min},\Delta m_J,\tau_{21,\min},\tau_{21,\max}).
\]
The resonant variable $m_{JJ}$ defines the candidate region but is not supplied
to the event localizer.  Unlike CATHODE, which learns the background
conditional on $m_{JJ}$ from sidebands, the present estimator study uses an
independent pure-QCD signal-region reference; this isolates localization and
Monte Carlo convergence from background-model learning.

The mixed mock sample contains $121352$ background and $772$ signal events in
the signal region. An independent evaluation sample contains $340000$
background and $20000$ signal events, and $272858$ additional signal-region
QCD events form the independent background reference. Each auxiliary variable
is mapped through its background CDF to copula coordinates.

To avoid committing the discrepancy measure to a single characteristic scale,
we use five Gaussian kernels. Taking the median pairwise distance between
background events in the transformed space as a reference scale,
$d_{\rm med}$, their bandwidths are
\[
h/d_{\rm med}=0.25,\,0.5,\,1,\,2,\,4.
\]
The smaller bandwidths are sensitive to relatively localized discrepancies,
while the larger ones probe broader departures. The resulting multiscale
kernel bank is fixed using background information only and is not optimized
using signal labels.

For computational efficiency, each Gaussian kernel is approximated by $512$
random Fourier features, as described in Sec.~\ref{sec:direct-estimation}. Concatenating the five
feature maps gives a total representation dimension of $5\times512=2560$.

For reference, the independent evaluation sample gives
\[
\AUC=0.91830,
\qquad
{\rm SIC}_{\max}=8.52,
\]
where ${\rm SIC}=\epsilon_S/\sqrt{\epsilon_B}$.  On the finite mixed mock
sample itself, the direct empirical witness---the convergence target for the
resampling estimators---gives
\begin{equation}
\AUC_{\rm witness}=0.916868,
\qquad
{\rm SIC}_{\max,\rm witness}=9.9495.
\label{eq:exp3-witness-target}
\end{equation}
The distinction is essential: the independent-sample numbers characterize
generalization performance, whereas Eq.~\eqref{eq:exp3-witness-target} is the
same-sample finite-data target whose ranking the stochastic estimators should
recover.

As a post-hoc check of the bandwidth coverage, we also evaluated single-scale
witnesses over the wider range
\[
h/h_{\rm med}=1/16,\,1/8,\,1/4,\,1/2,\,1,\,2,\,4,\,8,\,16.
\]
The independent-sample AUC reaches its maximum near $h/h_{\rm med}=1$ and
decreases toward smaller bandwidths, while extending the scan beyond
$h/h_{\rm med}=4$ produces no improvement. The original multiscale range
$0.25$--$4$ therefore brackets the region relevant for the overall event
ranking. Its combined AUC, $0.9183$, is close to the best single-scale value,
$0.9198$, without selecting a bandwidth using signal labels.

\paragraph{Historical raw resampling.}

Following our original motivation, here we first reproduce the global-batch Inverse-Bagging-style construction:
batches are drawn with replacement from the mixed mock sample, the
known-background unbiased MMD statistic is evaluated, and its value is assigned
to every event occurrence.  The fitted conditional slope agrees with the exact
$2/m$ expectation, confirming that the estimator is statistically correct, but
the ranking converges slowly.  At $m=1000$ and $R=5\times10^6$ global batches,
\[
\rho_{\rm raw}=0.7639,
\qquad
\AUC_{\rm raw}=0.8884,
\qquad
{\rm SIC}_{\max,\rm raw}=4.01.
\]
Here and below, $\rho$ denotes the correlation with the direct empirical
witness.  Raw
conditional averaging is therefore a correct but inefficient route to the
event localizer.

\paragraph{Marginal and event-containing estimators.}
For each event occurrence we also compute the matched marginal contribution and
the event-containing pair term.  At fixed $R=10^6$, increasing $m$ from $20$
to $1000$ raises the pair correlation with the direct witness from about
$0.227$ to $0.9966$, while ${\rm SIC}_{\max}$ rises from about $1.01$ to $8.52$.
The generic marginal estimator becomes numerically almost indistinguishable
from the pair estimator at large $m$.

\begin{figure}[t]
    \centering
    \includegraphics[width=\textwidth]{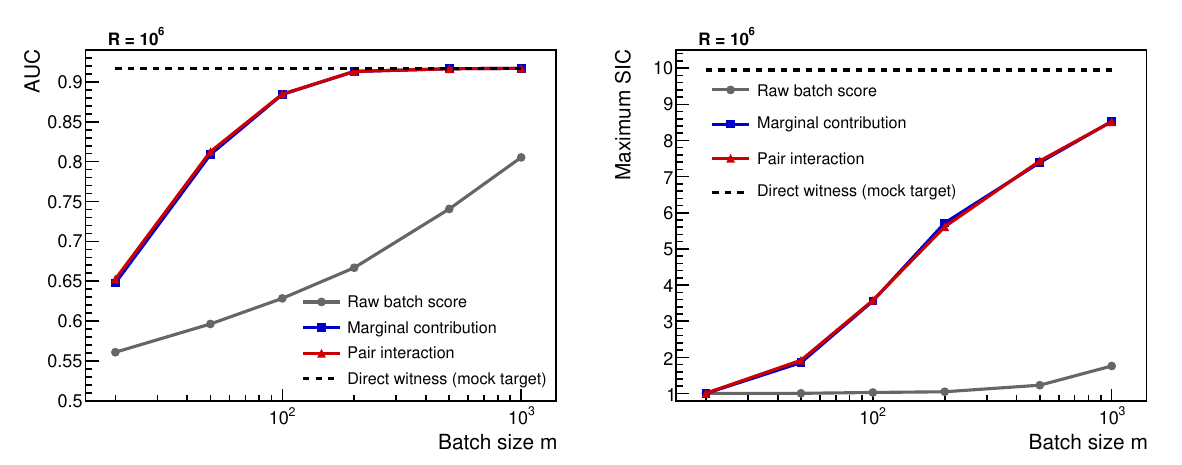}
    \caption{
    Raw conditional resampling, matched marginal contributions, and
    event-containing pair interactions versus batch size at fixed $R=10^6$ in
    the LHC Olympics mock sample.  \textit{Left:} AUC.  \textit{Right:}
    maximum SIC.  The horizontal line is the direct empirical witness on the
    same mock sample.  Increasing $m$ does not change the population ordering
    but strongly improves the Monte Carlo precision of marginal and pair
    estimators.
    }
    \label{fig:exp3-vs-m}
\end{figure}

At fixed $m=1000$, the pair estimator converges with $R$ as
\[
\begin{array}{c|cccccc}
R&10^5&2\times10^5&5\times10^5&10^6&2\times10^6&5\times10^6\\ \hline
\rho&0.9671&0.9831&0.9932&0.9966&0.99828&0.99931\\
{\rm SIC}_{\max}&6.60&7.43&8.38&8.52&8.79&9.19
\end{array}
\]
while the AUC is already saturated at the direct-witness value by about
$R=10^6$.  The slower convergence of SIC shows that correct ordering in the
extreme tail is substantially more demanding than bulk ROC ranking.

\begin{figure}[t]
    \centering
    \includegraphics[width=\textwidth]{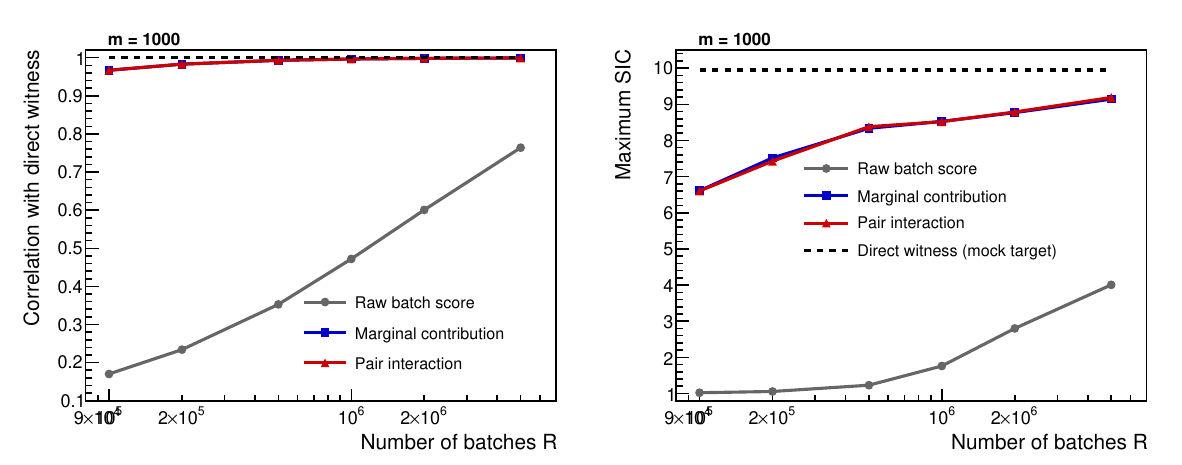}
    \caption{
    Convergence with the number of sampled batches at fixed $m=1000$.
    \textit{Left:} correlation with the direct empirical witness.
    \textit{Right:} maximum SIC.  Marginal and pair estimators rapidly
    approach the witness, whereas the raw batch score remains much noisier.
The substantially slower convergence of SIC than of the global score correlation shows that accurate ordering in the extreme tail is more demanding than reproducing the witness score field overall.
    }
    \label{fig:exp3-highm}
\end{figure}

At the largest run,
\begin{align}
\rho_{\rm marg}&=0.999304,
&\AUC_{\rm marg}&=0.916922,
&{\rm SIC}_{\max,\rm marg}&=9.143,\\
\rho_{\rm pair}&=0.999306,
&\AUC_{\rm pair}&=0.916915,
&{\rm SIC}_{\max,\rm pair}&=9.189.
\end{align}
Small AUC excursions above the direct-witness value at intermediate checkpoints
are finite-sample/Monte Carlo fluctuations and are not to be interpreted as superior
population discrimination.

\paragraph{Scaling law.}
The full scan provides a direct test of Eq.~\eqref{eq:q-rho-scaling}.  With
\[
q_\rho=\rho^{-2}-1,
\]
the pair-estimator points collapse onto a common inverse law in $Rm^2$ once the
sampling exposure is sufficiently large.  A log--log fit for
$Rm^2\geq2\times10^9$ to
\[
q_\rho=C(Rm^2)^{-\alpha}
\]
gives
\begin{equation}
\alpha=0.9995,
\qquad
C=6.80\times10^9,
\end{equation}
with an RMS scatter of $5.1\times10^{-3}$ in $\log q_\rho$ about the fitted
power law.
The fitted exponent is therefore remarkably close to the predicted $\alpha=1$; the lowest-exposure points show the expected pre-asymptotic deviation.

\begin{figure}[t]
    \centering
    \includegraphics[width=0.72\textwidth]{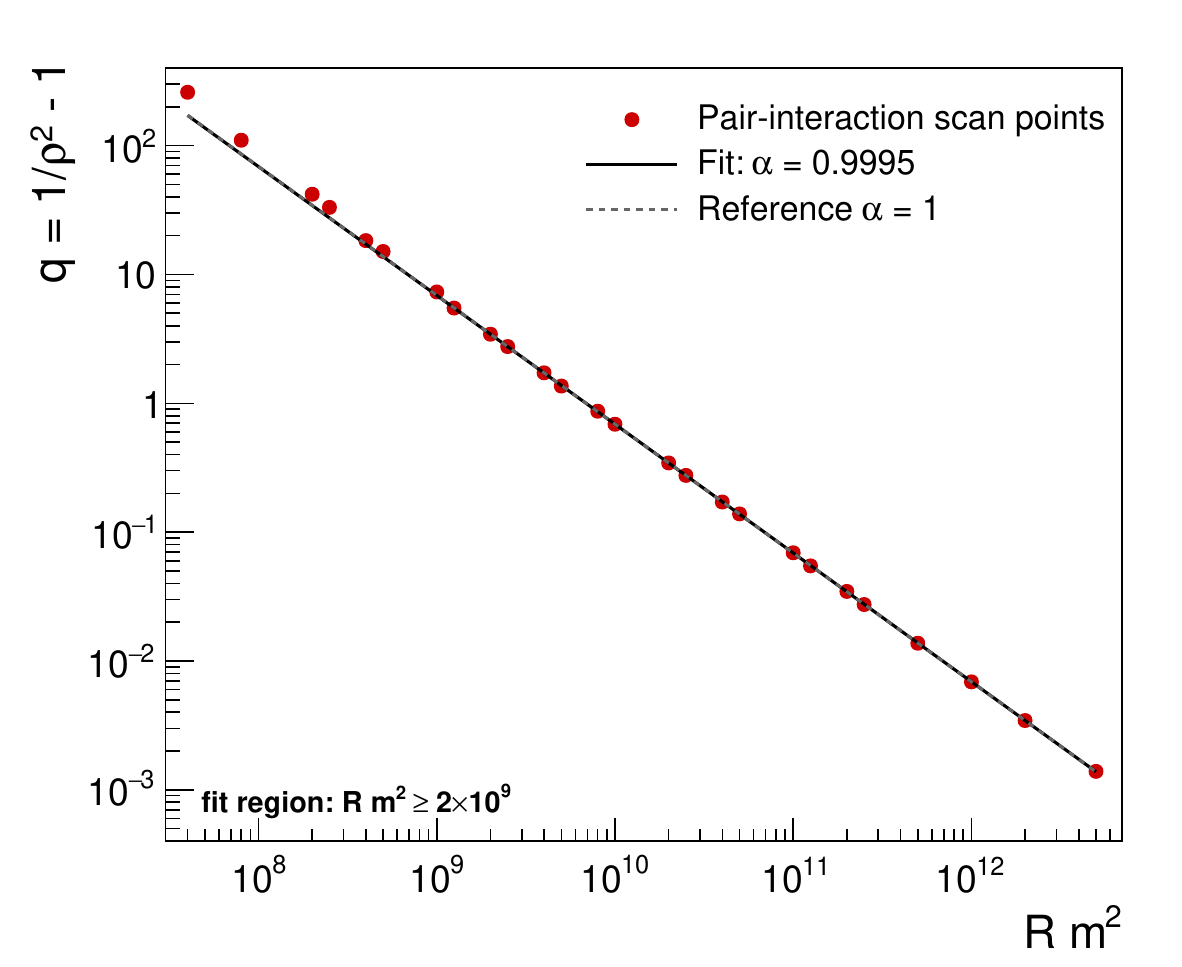}
    \caption{
    Scaling of the pair-interaction estimator.  The decorrelation measure
    $q_\rho=\rho^{-2}-1$ is plotted against $Rm^2$ for the different $m$ and
    $R$ configurations.  At sufficiently large exposure the points collapse
    onto a common power law.  The dashed line shows the predicted $\alpha=1$
    behavior.
    }
    \label{fig:exp3-scaling}
\end{figure}

Experiment~III therefore establishes the large-scale computational statement
of the paper.  Raw conditional resampling converges to the correct localizer
but slowly; matched marginal subtraction removes the dominant context
fluctuation; and for pairwise MMD the event-containing estimator realizes the
predicted $1/(Rm^2)$ stochastic scaling.  The near identity of the generic
marginal and MMD-specific pair estimators at large $m$ supports the former as
the broadly applicable construction and the latter as its transparent
U-statistic specialization.


\FloatBarrier
\section{Discussion}
\label{sec:discussion}

\subsection{A unified view of discrepancy localization}

The main conceptual result of this work is a common statistical interpretation
of constructions that appear separately in several established literatures.
Deletion and insertion diagnostics, marginal data contributions, conditional
subset averages, influence functions, and kernel witness functions were not
introduced for the same purpose, but they can all answer a closely related
question: how strongly is one observation associated with a discrepancy
exhibited by the sample as a whole?

For a symmetric statistic, the fixed-size replacement construction makes this
connection exact,
\[
A^{\rm rep}_{T,m}(x)=L_{T,m}(x),
\]
while for unbiased U-statistics the corresponding addition construction has
the same expectation. 

These constructions can be understood as different ways of assigning a
sample-level discrepancy back to individual observations: marginal
contributions and conditional projections identify the same local quantity,
while projection theory clarifies which low-order event contribution is being isolated.
For U-statistics the local quantity is
proportional to the first Hoeffding/H\'ajek projection; for smooth
functionals it agrees at leading order with the influence function; and for
the known-background MMD statistic it reduces exactly to the witness
function.  These component facts are classical or closely related to
established constructions.  The contribution of the present framework is to
place them in a common discrepancy-localization setting and to use that
identification to guide both interpretation and computation.
The extreme-background probe of Experiment I makes this distinction concrete: an observation far into the background tail has zero expected first-order MMD localization when embedded in genuine background contexts.

\subsection{Computational consequence: canceling irrelevant context}

The unified view has a direct algorithmic consequence.  Raw conditional
resampling estimates the correct local quantity, but every score retains the
fluctuation of the entire random context in which the event happened to be
placed.  A matched marginal difference removes the part of that fluctuation
that is common to the contexts with and without the event.  For U-statistics,
this cancellation can be pushed one step further by accumulating only terms
that actually contain the event being scored.

The numerical studies we have performed show that the above is not a cosmetic reformulation. Experiment~I verifies the expected variance reduction in a setting where the
population MMD localizer is analytically known.  In the near-degenerate regime
of that experiment the raw MMD statistic itself scales differently from the
generic nondegenerate first-order case, and the observed variance ratio follows
the corresponding $m^{-1}$ law rather than the generic $m^{-2}$ asymptote.
The event-containing estimator nevertheless gains the predicted additional
power of $m$ in its absolute variance.

Experiment~III allows us to perform a clean large-scale test.  Defining
$q_\rho=\rho^{-2}-1$, where $\rho$ is the correlation with the direct empirical MMD witness, the pair-interaction scan is described in the high-statistics regime by
\[
q_\rho=C(Rm^2)^{-\alpha},
\qquad
\alpha=0.9995,
\qquad
C=6.80\times10^9.
\]
The fitted exponent therefore agrees quantitatively with the predicted
$1/(Rm^2)$ behavior.  

At the largest sampled exposure the pair estimator reaches correlation 0.9993 with the direct witness and essentially saturates its AUC, while convergence in the extreme SIC tail remains slower.
The point is not that MMD requires resampling---its witness can be evaluated
directly---but that MMD provides an exact benchmark showing what is gained by
removing context-only fluctuations in a generic localization procedure.

\subsection{Localization is not the same as collective information}

A second distinction is conceptual rather than computational.  The existence
of a context-dependent estimator does not imply that an IID ensemble contains
additional class information about one member.  If signal and background
distributions are fully specified,
\[
\PP(Z_i=S\mid X_i,X_{-i})
=
\PP(Z_i=S\mid X_i).
\]
and no use of the remaining observations can improve the Bayes-optimal ranking
of event $i$.

The shared-latent construction isolates the genuinely different case.  Its
one-event signal marginal is deliberately identical to the background, so
$I(Z_i;X_i)=0$ and isolated classification has $\AUC=0.5$.  The remaining
events nevertheless constrain a parameter shared by the signal population,
which makes
\[
I(Z_i;X_{-i}\mid X_i)>0.
\]
Context has not created a new microscopic property of the event.  Rather, it has
learned which alternative is present and projected that information back onto
the individual observation.  Experiment~II demonstrates this distinction
numerically: the shared-$\mu$ ensemble yields a nontrivial contextual ranking,
whereas the independent-$\mu_i$ control has no systematic class information
above chance despite reproducing the finite-sample fluctuations of its own
empirical localizer.
As the ensemble size increases, this is not merely a qualitative effect: at N=20000 the empirical contextual witness recovers about 94\% of the MMD classification advantage available when the shared latent alternative is known.

This distinction also clarifies the role of Experiment~III.  There the main
question is efficient estimation of a localizer defined by a fixed mixed
sample.  
Experiment II, rather than Experiment III, provides the numerical demonstration of the stronger information-theoretic case established in Sec. 5: the surrounding ensemble can contain class information absent from the isolated observation.

\subsection{Inverse Bagging and collider anomaly searches}

Inverse Bagging provides the historical anomaly-detection instance that
motivated the present analysis.  Its original implementation bundled bootstrap
sampling with replacement, small batches, a goodness-of-fit statistic,
optional thresholding, and aggregation of batch properties back onto the
events that participated in them.  The present framework separates those
choices.  The aggregation step is a Monte Carlo estimate of conditional
localization; matched marginal subtraction estimates the same population
target more efficiently; and event-containing U-statistic terms expose the
relevant projection directly when the parent statistic permits it.  In that
sense the framework provides both a statistical interpretation of Inverse
Bagging and a route beyond its particular bootstrap implementation.

For modern collider searches, the resulting task is complementary to ordinary
outlier detection and density-ratio learning.  Methods such as CATHODE
\cite{hallin2022}, NPLM-like approaches \cite{dagnolo2019}, classifier-based
two-sample tests, and kernel discrepancies optimize different statistical
objectives.  The LHC Olympics study in Sec.~\ref{sec:exp-lhco} is therefore not
a claim of universal superiority over contemporary anomaly detectors.  It
shows instead that a sample-level nonparametric discrepancy can be localized
efficiently at collider scale.

An attractive feature of this viewpoint is that the meaning of
``responsibility for the anomaly'' is inherited from the parent discrepancy.
Different global statistics will induce different local rankings because they
respond to different departures from the reference model.  Localization is
therefore interpretable relative to an explicit sample-level statistical
question rather than to a universal notion of outlierness.

\subsection{Limitations and statistical inference}

Several limitations remain to be investigated in future work.  The background model may contain nuisance
parameters or be learned from the same data; observations may not be IID;
kernel or feature choices may be data-adaptive; and a discrepancy may be
intrinsically higher-order, with a vanishing first projection and therefore
requiring pair- or group-level localization.  Extending the framework to these
settings is more important than refining the already well-controlled MMD
examples considered here.

The localized scores are descriptive measures of association with a
collectively observed departure, rather than posterior signal
probabilities.  A global discovery claim and an event-level localization claim are separate statistical operations, as discussed in
Sec.~\ref{sec:global-local}.  If selected high-score events are subsequently
used for formal inference, data splitting or calibration of the complete
adaptive procedure is required.  Cross-fitting controls self-use in score
construction but does not by itself correct post-selection inference.

Finally, explicit resampling is not always the computationally preferred
implementation for the constructions discussed in this work.  When the localizer is analytically available, as for MMD, direct witness evaluation or random-feature approximations may be cheaper. The value of the marginal formulation is its generality: it gives a matched
estimator when no closed-form localizer is available and makes explicit which parts of the random context are irrelevant to the observation being scored.

\section{Conclusions}
\label{sec:conclusion}

We have developed a unified view of event-level localization for sample-level
discrepancy statistics.  Conditional subset averaging and marginal
contributions, familiar in different forms from resampling diagnostics and
data valuation, coincide under fixed-size replacement; for U-statistics the
resulting local quantity is the first Hoeffding/H\'ajek projection, while for
smooth discrepancies it connects to influence functions and for
known-background MMD it is exactly the witness function.  The contribution is
therefore not a new notion of marginal attribution, but the identification of
these constructions as closely related ways of projecting a global departure
onto individual observations.

That identification has a practical consequence.  Raw conditional resampling
retains fluctuations of the random context that do not involve the event being
scored.  Matched marginal differences cancel those fluctuations, and for
U-statistics event-containing terms isolate the relevant contribution even
more directly.  In the LHC Olympics study the pair estimator follows
\[
q_\rho=C(Rm^2)^{-\alpha},
\qquad \alpha=0.9995,
\]
in the asymptotic regime and reaches correlation $0.9993$ with the direct
empirical MMD witness.  This provides a quantitative demonstration that the
same event-level target can be recovered far more efficiently than by raw
batch averaging.

A separate question is whether the ensemble merely helps estimate an
ensemble-defined localizer or contains genuinely new information about the
class of an individual event.  For fully specified IID alternatives it does
not.  When anomalous observations share unknown latent structure, however,
the other observations can constrain that structure and thereby change the
posterior interpretation of each member.  Experiment~II demonstrates the
limiting case in which $I(Z_i;X_i)=0$ but
$I(Z_i;X_{-i}\mid X_i)>0$.

Taken together, these results provide a statistical bridge from global
discrepancy detection to interpretable event-level localization.  The bridge
clarifies what quantity is being attributed to an observation, how that
quantity can be estimated efficiently, and when the surrounding ensemble can
supply information that no event carries in isolation.  Natural extensions
include higher-order localizers for degenerate discrepancies, nuisance-aware
or learned reference models, and non-IID data.

\end{document}